\documentclass[entropy,article,accept,moreauthors,pdftex]{Definitions/mdpi} 

\pdfoutput=1
\firstpage{1} 
\pubvolume{21}
\issuenum{11}
\articlenumber{1109}
\pubyear{2019}
\copyrightyear{2019}
\history{Received: 30 September 2019; Accepted: 4 October 2019; Published: 12 November 2019}
\updates{yes} 

\usepackage{amssymb, bm}
\usepackage{algorithm}
\usepackage{algpseudocode}
\usepackage{dcolumn}
\usepackage{ifdraft}
\usepackage{hyperref}
\hypersetup{%
	bookmarksopen={true},
	pdffitwindow=true, 
	colorlinks=true, 
	linkcolor=bluecite, 
	citecolor=bluecite, 
	urlcolor=bluecite, 
	hyperfootnotes=false, 
	pdfstartview={FitH},
	pdfpagemode= UseNone
}

\newcommand{\ubw}{{\mathbf{w}}}
\newcommand{\uby}{{\mathbf{y}}}
\newcommand{\ubx}{{\mathbf{x}}}

\newcommand{\ubv}{{\mathbf{v}}}
\newcommand{\ubn}{{\mathbf{n}}}
\newcommand{\bmtheta}{{\bm{\theta}}}
\newcommand{\data}{{\mathcal{D}}}
\newcommand{\ev}{{\mathcal{Z}}}

\DeclareMathOperator*{\argmin}{argmin}
\DeclareMathOperator*{\ESS}{ESS}

 \usepackage[labelformat=simple]{subcaption}

\DeclareCaptionLabelFormat{subcaptionlabel}{\normalfont(\textbf{#2}\normalfont)}
\Title{
  Stochastic Gradient Annealed Importance Sampling
  for Efficient Online Marginal Likelihood Estimation $^\dagger$
}

\Author{Scott A. Cameron $^{1,2,}$*\orcidA{}, Hans C. Eggers $^{1,2}$\orcidB{} and Steve Kroon $^{3}$\orcidC{}}

\AuthorNames{Scott A. Cameron, Hans C. Eggers and Steve Kroon}

\address{%
$^{1}$ \quad Department of Physics, Stellenbosch University, Stellenbosch 7600, South Africa; eggers@sun.ac.za\\
$^{2}$ \quad National Institute for Theoretical Physics, Stellenbosch 7600, South
Africa\\
$^{3}$ \quad Computer Science Division, Stellenbosch University, Stellenbosch 7600, South Africa; kroon@sun.ac.za}

\corres{Correspondence: scott.a.cameron@live.co.uk}

\conference{MaxEnt~2019} 

\abstract{
We consider estimating the marginal likelihood in settings with independent and identically distributed (i.i.d.) data.
We propose estimating the predictive distributions in a sequential factorization
of the marginal likelihood in such settings by using stochastic gradient Markov
Chain Monte Carlo techniques. 
This approach is far more efficient than traditional marginal likelihood
estimation techniques such as nested sampling and annealed importance sampling
due to its use of mini-batches to approximate the likelihood. 
Stability of the estimates is provided by an adaptive annealing schedule.
The resulting stochastic gradient annealed importance sampling~(SGAIS)
technique,
which is the key contribution of our paper,
enables us to estimate the marginal likelihood of a number of models
considerably faster than traditional approaches, with no noticeable loss of
accuracy.
An important benefit of our approach is that the marginal likelihood is
calculated in an online fashion as data becomes available, allowing the
estimates to be used for applications such as online weighted model combination.
}

\keyword{marginal likelihood; evidence; nested sampling; annealed importance
sampling; Monte Carlo; stochastic gradients; SGHMC}

\begin{document}
\section{Introduction}

Marginal likelihood (ML), sometimes called evidence, is a quantitative measure of
how well a model can describe a particular data set; it is the probability
that
the data set occurred within that model.  Consider a Bayesian model with
parameters $\bmtheta$ for a data set $\data = \{\uby_n\}_{n=1}^{N}$.  The
ML is the integral
\begin{equation*}
  \ev := p(\data) = \int p(\data|\bmtheta) p(\bmtheta) \, d\bmtheta,
\end{equation*}
where $p(\data|\bmtheta)$ is the likelihood and $p(\bmtheta)$ is the prior. In
this paper, we consider the case where the data are conditionally independent
given the parameters
{$p(\data|\bmtheta) = \prod_{n}
p(\uby_n|\bmtheta)$},
as is common in many parametric models. This restriction is to ensure
that a central limit theorem applies to the stochastic likelihood approximation.
This can be weakened to any factorization that
exhibits a central limit theorem, such as conditionally Markov data and
autoregressive models. 
The~posterior distribution over a set of models is proportional to their
MLs, and so approximations to ML are sometimes used for model comparison and
weighted model averaging~\mbox{\cite{bayesian-reasoning} (chapter 12)}.
The above integral is typically analytically intractable for any but the
simplest models, so one must resort to numerical approximation methods.

Nested sampling~(NS)~\cite{nested-sampling} and annealed importance
sampling~(AIS)~\cite{ais} are two algorithms able to produce accurate estimates
of the ML. NS accomplishes this by sampling from the prior under constraints of
increasing likelihood and AIS by sampling from a temperature annealed
distribution~$\propto~p(\data|\bmtheta)^{\lambda}p(\bmtheta)$ and averaging over
samples with appropriately calculated importance weights.
Although NS and AIS produce accurate estimates of the ML, they
tend to scale poorly to large data sets due to the fact that they need to
repeatedly calculate the likelihood function: for NS, this is to ensure staying
within the constrained likelihood contour; for AIS, the likelihood must be
calculated both to sample from the annealed distributions using some Markov
chain Monte Carlo (MCMC) method as well as to calculate the importance
weights.
Calculation of the likelihood is computationally expensive on large data sets.
To combat this problem, various optimization and sampling algorithms instead 
use stochastic approximations of the likelihood by sub-sampling the data set
into mini-batches~\cite{sgld}.

Unfortunately, NS and AIS cannot trivially use mini-batching in their vanilla
form to improve
scalability. Using stochastic likelihood approximations changes the statistics
of the likelihood contours in NS, allowing particles to occasionally move to
lower likelihood instead of higher, violating the basic assumptions of the
algorithm. Vanilla AIS could benefit from using stochastic likelihood
gradients during
the MCMC steps, but introducing stochasticity into the importance weights would
bias the results.

The key contributions of this work are as follows:
\begin{itemize}[leftmargin=*,labelsep=5.8mm]
  \item
      We introduce stochastic gradient annealed importance sampling~(SGAIS),
      which combines stochastic gradient MCMC with annealed
      importance sampling to estimate the ML in an online fashion using
      mini-batch Bayesian updating.
  \item 
      SGAIS enables efficient ML estimation for streaming data and for large
      data sets, which was not previously feasible.
  \item 
      We illustrate how SGAIS can be used to identify distribution shift in the
      data when applied in an online setting.
  \item 
      We empirically analyze the behavior of SGAIS and its robustness to
      various choices of algorithm parameters.
\end{itemize}

We illustrate our approach by calculating
ML estimates on simulated data generated with three simple models.
For these models, we obtain considerable speedup over nested
sampling and annealed importance sampling 
on data sets with one million observations, without noticeable loss in
accuracy.

\section{Sequential Marginal Likelihood Estimation}
\label{sec:seq}

The ML can be factorized, through the product rule, into a
product of predictive distributions
{$\ev = \prod_{n} p(\uby_n|\uby_{<n})$},
where
\begin{equation}
  \label{eq:pred}
  p(\uby_n|\uby_{<n}) = \int p(\uby_n|\bmtheta) p(\bmtheta|\uby_{<n}) \,
  d\bmtheta.
\end{equation}
Throughout this manuscript, we will present our approach as Bayesian updating
based on one observation at a time. 
We do this for notational simplicity, with the understanding that the extension
to Bayesian updating with multiple observations is mathematically trivial. 
In our experiments, Section~\ref{sec:experiments}, we decompose the data into
chunks of a fixed size, rather than one data point at a time.
Assuming one is able to produce accurate estimates $\hat{p}(\uby_n|\uby_{<n})$
of the predictive probabilities, the log-ML can be approximated by
$\log\hat\ev = \sum_n \log\hat{p}(\uby_n|\uby_{<n})$.
In this way, the difficult problem of estimating an integral of an extremely
peaked function, $p(\data|\bmtheta)$, reduces to the easier problem of
estimating many integrals of smoother functions $p(\uby_n|\bmtheta)$.

Many sequential Monte Carlo (SMC) methods use a similar approach and
calculate predictive estimates using a combination of importance resampling and
MCMC mutation steps~\cite{smc}. Generic examples of such algorithms are the
bootstrap particle filter~\cite{bootstrap-filter}, which is often used for
posterior inference in hidden Markov models and other latent variable sequence
models~\cite{smc}, and the ``left-to-right'' algorithm, which is used
in~\cite{topic-models} to evaluate topic models.

The computational efficiency of using this approach depends on the method of
approximating Equation~(\ref{eq:pred}).
In previous work~\cite{maxent-paper}, we proposed the estimator
$\hat{p}(\uby_n|\uby_{<n}) = \frac{1}{M}
\sum_{i=1}^M p(\uby_n|\bmtheta_i)$,
where each $\bmtheta_i$ is drawn from the posterior distribution
$p(\bmtheta|\uby_{<n})$ using MCMC methods. 
This paper extends this idea by combining Bayesian updating
with annealing, which we describe in Section~\ref{sec:annealing}.
Since samples from the previous posterior, $p(\bmtheta|\uby_{<n-1})$, would
generally be available at each step, we expect that only a small number of steps will
be needed to accurately sample from the next posterior distribution,
$p(\bmtheta|\uby_{<n})$.
Metropolis--Hastings-based MCMC algorithms would have to iterate over all
previous $n-1$ data points in order to calculate the acceptance probability
for each Markov transition, and so using them to estimate $\log\ev$ in this
sequential manner would scale at least quadratically in $N$.
The key computational improvement in our approach comprises the use of stochastic gradient-based MCMC algorithms such as the stochastic gradient Hamiltonian Monte
Carlo (SGHMC)~\cite{sghmc}. SGHMC utilizes mini-batching, allowing one to efficiently
draw samples from the posterior distribution $p(\bmtheta|\uby_{<n})$ even when
$n$ is large. We call this approach stochastic gradient annealed importance
sampling~(SGAIS).

\section{Bayesian Updating with Annealing}
\label{sec:annealing}

Annealed importance sampling~(AIS)~\cite{ais} is a Monte Carlo method that
generates samples from a sequence
of intermediate distributions, bridging from a simple distribution that can be
sampled easily (usually the prior) to a desired distribution (usually the
posterior). 
AIS also produces an unbiased estimate of the
normalizing constant (the ML) for the desired distribution as a byproduct.
The sequence of distributions is typically chosen to have the form
{$f_t(\bmtheta) = p(\data|\bmtheta)^{\lambda_t}
p(\bmtheta)$},
where 0 = $\lambda_0 < \lambda_1 < \ldots < \lambda_T = 1$. 
At time step $t=0$, a number of particles are drawn from the prior, and at each
subsequent time step $t = 1,\ldots,T$, an MCMC operator that leaves
the distribution $f_t$ invariant is applied to each particle. 
The importance weights are initialized as the
normalizing constant of the prior, which is typically 1, and at each time step
are updated as follows:
\begin{equation*}
  w_i^{(t)} \gets w_i^{(t-1)} \frac{f_t(\bmtheta_i^{(t-1)})}{f_{t-1}(\bmtheta_i^{(t-1)})} = 
  w_i^{(t-1)} \, p(\data|\bmtheta_i^{(t-1)})^{\lambda_{t} - \lambda_{t-1}}
  \qquad i = 1,\ldots,M,
\end{equation*}
where $\bmtheta_i^{(t)}$ is the $i$\textsuperscript{th} particle at time step
$t$.
For any function $h$, the unnormalized posterior expectation
\begin{equation*}
  \int h(\bmtheta) f_T(\bmtheta) \, d\bmtheta = \int h(\bmtheta)
  p(\data|\bmtheta)p(\bmtheta) \, d\bmtheta
\end{equation*}
can be estimated using $M$ particles by 
$\frac{1}{M} \sum_i h(\bmtheta^{(T-1)}_i) w^{(T)}_i$.
This
estimator $\hat{H}$ is unbiased and corresponds to the ML in the case
$h(\bmtheta) = 1$ and the product of the ML and the posterior predictive
$p(\uby'|\data)$ in the case $h(\bmtheta) = p(\uby'|\bmtheta)$.
Although it is possible to use stochastic gradient-based MCMC algorithms to
update the particles in this setting, the importance weight updates would
normally still require iterating over the whole data set at each time step.
To circumvent this, we propose using AIS sequentially to calculate
predictive probabilities in a Bayesian updating
setting, as described in Section~\ref{sec:seq}, instead of on the full data
set.
This is equivalent to
using the following sequence of intermediate distributions:
\begin{equation}
  \label{eq:sais-dist}
  f_n^{(t)}(\bmtheta) =
  p(\uby_n|\bmtheta)^{\lambda_t}
  \left[\prod_{k<n}p(\uby_k|\bmtheta)\right] p(\bmtheta).
\end{equation}
In principle, $\lambda$ may now depend on both $n$ and $t$. Since we will choose the
annealing schedule adaptively anyway, this notation is suppressed.
From an SMC perspective, this can be considered a combination of thermal and data
tempering~\cite{smc}.
The corresponding importance weights can then be calculated without iterating
over the entire data set;
\begin{equation*}
  w_i^{(t)} \gets w_i^{(t-1)}
  \frac{f_n^{(t)}(\bmtheta_i^{(t-1)})}{f_n^{(t-1)}(\bmtheta_i^{(t-1)})}
  = w_i^{(t-1)} p(\uby_n|\bmtheta_i^{(t-1)})^{\lambda_t - \lambda_{t-1}}.
\end{equation*}
This result is of central importance to this paper: only by using AIS within
a Bayesian updating setting can we effectively take advantage of
mini-batching and stochastic gradients.

So far,
the choice of each $\lambda_t$ is not specified.
While any increasing sequence of values for $\lambda_t$ (annealing schedule)
guarantees an unbiased estimator of the ML, a poor choice can result in high
variance of the ML estimator. 
Grosse~et~al.~\cite{sandwich} point out that a linear sequence typically
results in poor performance, and instead, they recommend a sigmoidal annealing
schedule, $\lambda_t = \sigma(\delta(2t/T - 1))$ for some $\delta$. 
We expect that during early Bayesian updating steps, i.e., when $n$ is small, the relative prior
$p(\bmtheta|\uby_{<n-1})$ and posterior $p(\bmtheta|\uby_{<n})$ may differ
substantially, and so many intermediate distributions may be needed. However,
once $n$ becomes large, the relative prior and posterior will likely be similar
and only a few intermediate distributions should be required.
For this reason, it is preferable that the annealing schedule be chosen
adaptively.
An adaptive annealing schedule can be chosen by specifying a target effective
sample size~(ESS)~\cite{ess} and choosing the next $\lambda$ such that the ESS
is approximately equal to this
target~\cite{adaptive-smc,adaptive-smc-convergence}. 
For the choice of intermediate distributions given in Equation~(\ref{eq:sais-dist}), the
ESS used for the weight update is
\begin{equation*}
  \ESS(\Delta) := \frac{(\sum_i \omega_i(\Delta))^2}{\sum_i \omega_i(\Delta)^2},
\end{equation*}
where $\Delta := \lambda_t - \lambda_{t-1}$ and
\begin{equation*}
  \omega_i(\Delta) :=
  \frac{f_n^{(t)}(\bmtheta_i^{(t-1)})}{f_n^{(t-1)}(\bmtheta_i^{(t-1)})} =
  p(\uby_n|\bmtheta_i^{(t-1)})^{\Delta}.
\end{equation*}

Our proposed method combining sequential Bayesian updating and AIS 
is outlined in Algorithm~\ref{alg:sais}. As in other SMC
algorithms, one can optionally resample the particles $\bmtheta_i$
proportionally to their importance weights $w_i$ before performing each MCMC
update on line~\ref{alg:sais-sghmc}, thereafter setting
$w_i~\gets~\frac{1}{M}\sum_i w_i$.
Resampling can help to reduce variance in the estimator but can cause mode
collapse if few particles are used.
See~\cite{smc,adaptive-smc} for more details on resampling.
After each annealing sequence is completed (i.e., after the completion
of the inner while-loop in the algorithm), one can estimate the ML of the data
observed so far by $\hat{p}(\uby_{\le n}) := \frac{1}{M} \sum_i w_i$.

\begin{algorithm}[H]
  \caption{Stochastic Gradient Annealed Importance Sampling}
  \label{alg:sais}
 \normalsize{
 \setlength\baselineskip{15pt}
  \textbf{Input:} Data $\data=\{\uby_n\}_{n=1}^N$, number of particles $M$,
  pdfs $p(\uby|\bmtheta),p(\bmtheta)$, target ESS: $\ESS^*$ \\
  \textbf{Output:} $\hat{\ev}$ estimator of marginal likelihood
  \begin{algorithmic}[1]
    \State $\forall_i$: sample $\bmtheta_i \sim p(\bmtheta)$
    \State $\forall_i$: $w_i \gets 1$
    \For{$n = 1,\ldots,N$}
      \State $\lambda \gets 0$
      \While{$\lambda < 1$} 
        \State $\Delta \gets \argmin_{\Delta}[\ESS(\Delta) - \ESS^*]$
        \Comment{$\Delta \in (0, 1-\lambda]$}
        \State $\lambda \gets \lambda + \Delta$
        \State $\forall_i$: $w_i \gets w_i p(\uby_n|\bmtheta_i)^\Delta$
        \State $\forall_i$: $\bmtheta_i \gets \Call{Sghmc}{\bmtheta_i, \hat{U}_n^{(\lambda)}}$ 
        \Comment{potential energy $\hat{U}_n^{(\lambda)}$ defined in
        Equation~(\ref{eq:seq-potential})}
        \label{alg:sais-sghmc}
      \EndWhile \label{alg:sais-end-while}
    \EndFor
    \State \textbf{return} $\hat{\ev} = \frac{1}{M}\sum_i w_i$
  \end{algorithmic}}
\end{algorithm}

\section{Stochastic Gradient Hamiltonian Monte Carlo}
\label{sec:sghmc}

SGHMC~\cite{sghmc} generically simulates a Brownian particle in a potential
$U(\bmtheta)$ by
numerically integrating the Langevin equation
\begin{equation*}
  \begin{array}{ll}
    d\bmtheta &= \ubv\, dt,\\
    d\ubv &= - \nabla U(\bmtheta)\, dt - \gamma \ubv\, dt + \sqrt{2\gamma}\, dW,
  \end{array}
\end{equation*}
where $\gamma$ is the friction
coefficient and $W$ is the standard Wiener process.
See~\cite{stochastic-calculus} (chapter 5) for an introduction to stochastic
differential equations.
It can be shown through the use of a Fokker--Planck equation~\cite{gardiner} that
the above dynamics converge to the stationary distribution
$p(\bmtheta, \ubv) \propto \exp(- U(\bmtheta) - \frac{1}{2}\ubv^2)$.
We can use this to sample from the full data posterior by using a potential
energy equal to the negative log joint, $U(\bmtheta) := -\log{p}(\data,
\bmtheta)$.
The numeric integration is typically discretized~\cite{sghmc, cyclic-sghmc} as
\begin{equation}
  \label{eq:sghmc}
  \begin{array}{ll}
    \Delta \bmtheta &= \ubv, \\
    \Delta \ubv &= - \eta \nabla \hat{U}(\bmtheta) - \alpha \ubv +
    \bm{\epsilon} \sqrt{2(\alpha - \hat{\beta})\eta},
  \end{array}
\end{equation}
where $\eta$ is called the learning rate, $1 - \alpha$ is the momentum
decay, $\bm{\epsilon}$ is a standard Gaussian random vector,
$\hat\beta$ is an optional parameter to offset the variance of the stochastic
gradient term, and $\hat{U}(\bmtheta)$
is an unbiased estimate of $U(\bmtheta)$ calculated on independently
sampled mini-batches.
Since $U(\bmtheta)$ grows with the size of the data set, a small learning rate
$\eta \sim \mathcal{O}\left(\frac{1}{|\data|}\right)$ is required to minimize
the discretization  error.
The variance of the stochastic gradient term is proportional to $\eta^2$, while
the variance of the injected noise is proportional to $\eta$, so in the limit
$\eta \to 0$, stochasticity in the gradient estimates becomes negligible and the
correct continuum dynamics are recovered, even if one ignores the errors from
the stochastic gradient noise by using $\hat{\beta} = 0$. We refer the reader
to~\cite{sgrhmc,bohamiann,sghmc} for an in-depth analysis of the algorithm
parameters.

For the purposes of Bayesian updating and annealing, we use SGHMC with a
potential energy that leaves the distribution $f_n^{(t)}$ in
Equation~(\ref{eq:sais-dist}) invariant, $U(\bmtheta) = - \log f_n^{(t)}(\bmtheta)$.
We can approximate this potential energy stochastically by
\begin{equation}
  \label{eq:seq-potential}
  \hat{U}_n^{(\lambda)}(\bmtheta) =
  -\lambda\log{p}(\uby_n|\bmtheta)
  -\frac{n-1}{|B|}\sum_{\uby \in B} \log{p}(\uby|\bmtheta)
  -\log{p(\bmtheta)},
\end{equation}
where the mini-batch is drawn independently and identically distributed (i.i.d.) 
with replacement from the set of all
previous data points, i.e., $B \subset \{\uby_k|k < n\}$. Since
$\hat{U}_n^{(\lambda)}(\bmtheta)$ is an unbiased estimator of $-\log
f_n^{(t)}(\bmtheta)$ with finite variance (assuming the model assigns a non-zero probability to all observations), 
SGHMC with this potential energy will
leave $f_n^{(t)}$ invariant in the limit $\eta \to 0$.

\section{Online Marginal Likelihood Estimation}
\label{sec:online}

Our proposed approach is particularly efficient for ML estimation in
applications
in which new data are continually becoming available, since the marginal cost to
update the ML estimate based on new data is independent of the amount of
previously processed data.
Some examples of such applications are: finance modeling, monitoring audio or
video or other sensor readings, applications in security, process control, etc.
In these types of applications, vanilla NS or AIS would have to recalculate the
ML from scratch, without leveraging the previous ML estimates.

One challenge with estimating ML online using SGAIS is the generation of
mini-batches that are uniformly sampled from historical data.
This challenge can be tackled with the help of reservoir sampling.
If a large enough reservoir is kept to be representative of the previously
processed data, mini-batches could be drawn uniformly from the reservoir
instead~\cite{reservoir-sampling}.
If the data set is too large to fit on disk or is only available during
streaming, then this approach makes it possible to estimate ML, whereas using NS
or AIS to estimate ML would not even be possible.

In online scenarios, the data may also exhibit various degrees of
non-stationarity.
This may happen due to a shift in the underlying data-generating distribution.
Such distribution shifts would typically cause a noticeable change in the ML,
which can then be used for change-point detection.

\section{Experiments\label{sec:experiments}}
\unskip
\subsection{Experimental Setup}
To evaluate the accuracy and run-time performance of our proposed approach, we
estimate the ML for three simple models on simulated data sets in an online
fashion.
We further evaluate the robustness of SGAIS under various choices of algorithm
parameter values.

\textbf{Default Parameters.}
We use mini-batches of size 500, with the following SGHMC
parameters:
$\eta =
0.1/N$, $\alpha = 0.2$, and $\hat{\beta} = 0$. Predictive distributions are
approximated using $M = 10$ particles and 20 burn-in steps for each intermediate
distribution.
We use a target $\ESS$ of 5 for adaptive annealing.
As~mentioned in Section~\ref{sec:seq}, rather than Bayesian updating by adding a
single observation at a time, we add chunks of data at a time that are the same
size as the mini-batches.
These parameter choices are not necessarily the optimal choice for the models
below.
Instead, we chose the SGHMC parameters based approximately on those suggested
in~\cite{sghmc}, and we chose the number of particles and target $\ESS$ to
hopefully be sufficient for adaptive annealing but small enough to result
in a short running time.

\textbf{NS and AIS.}
As our reference standards of accuracy, we implemented NS and AIS.
Both NS and AIS implementations use SGHMC as their MCMC kernel.
For NS, this is to sample from the prior; for AIS, this is to sample from the
intermediate annealed distributions.
For AIS, we use the same parameters as for our sequential sampler, except that
each MCMC step uses the whole data set instead of mini-batches.
We implement NS with 20 SGHMC steps to sample from the constrained prior.
NS still requires the full data set to check the constraints and so cannot take
advantage of mini-batching.
For SGHMC used with NS, we used
parameters $\eta = 10^{-3}$, $\alpha = 0.1$, and $\hat{\beta} = 0$
because there is no gradient noise when sampling from the prior. Results
reported are for two particles; more behave similarly but are slower.  We allow
NS to run until the additive terms are less than 1\% of the current $\hat\ev$
estimate. This is a popular stopping criterion and is also used
in~\cite{sandwich}. 
More information about our implementations is given in Appendix~\ref{ap:ns}.

\textbf{Sensitivity Analysis.}
We investigate the sensitivity of our approach to the following parameters:
number of particles $M$, the target $\ESS$, the number of burn-in steps for each
intermediate distribution, the learning rate $\eta$, and the mini-batch size.
Each test is done by varying one parameter while keeping the others fixed at
their default values.

\textbf{Run-time Environment.}
Our experiments were executed on a laptop with an Intel i7 CPU and 8GB of ram,
running Linux. 
For fair comparison, all code was single-threaded. Multithreading gives a
considerable speedup when calculating the likelihood on large data sets but can
introduce subtle complexities that are difficult to control for in tests of
run-time performance.

\textbf{Models.}
We evaluated our approach on 3 models:
\begin{itemize}[leftmargin=*,labelsep=5.8mm]
  \item Bayesian linear regression with 6 parameters.
  \item 
      Bayesian logistic regression with 10-dimensional observations and four
      classes. This~model has 44 parameters.%
  \item 
      Bayesian Gaussian mixture model for two-dimensional data with 5 mixture
      components. This~model has 25 parameters.%
\end{itemize}
For each of the models, we generated data sets by sampling i.i.d.
from the model's conditional distribution with the parameters fixed.
See Appendix~\ref{ap:models} for a more detailed description of the models.%

\subsection{Distribution Shift}
\label{sec:exp-dist-shift}

\begin{figure}[H]
  \begin{subfigure}{0.24\textwidth}
    \includegraphics[width=\textwidth]{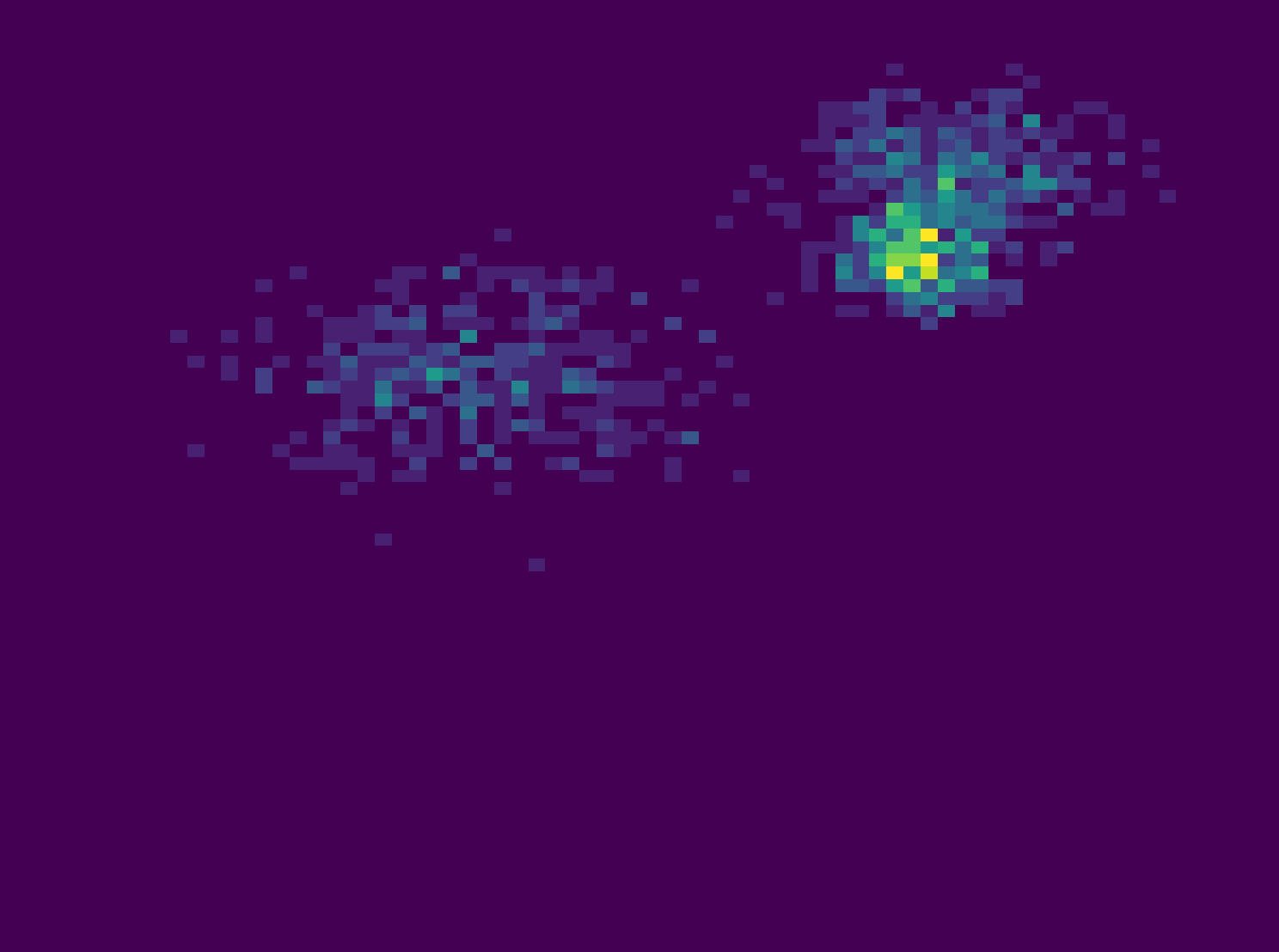}
    \caption{}
    \label{fig:dist-1k}
  \end{subfigure}
  \begin{subfigure}{0.24\textwidth}
    \includegraphics[width=\textwidth]{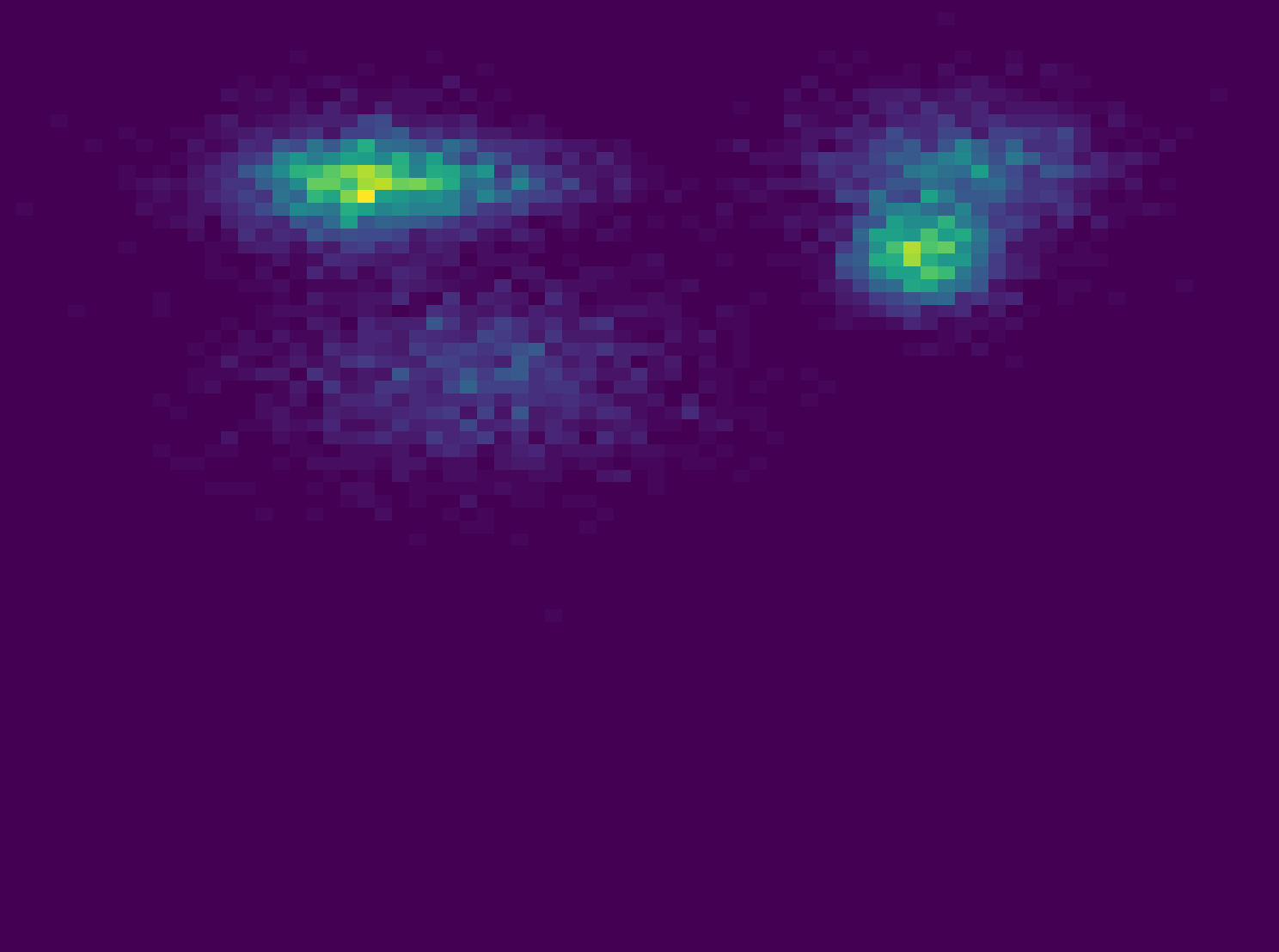}
    \caption{}
    \label{fig:dist-9k}
  \end{subfigure}
  \begin{subfigure}{0.24\textwidth}
    \includegraphics[width=\textwidth]{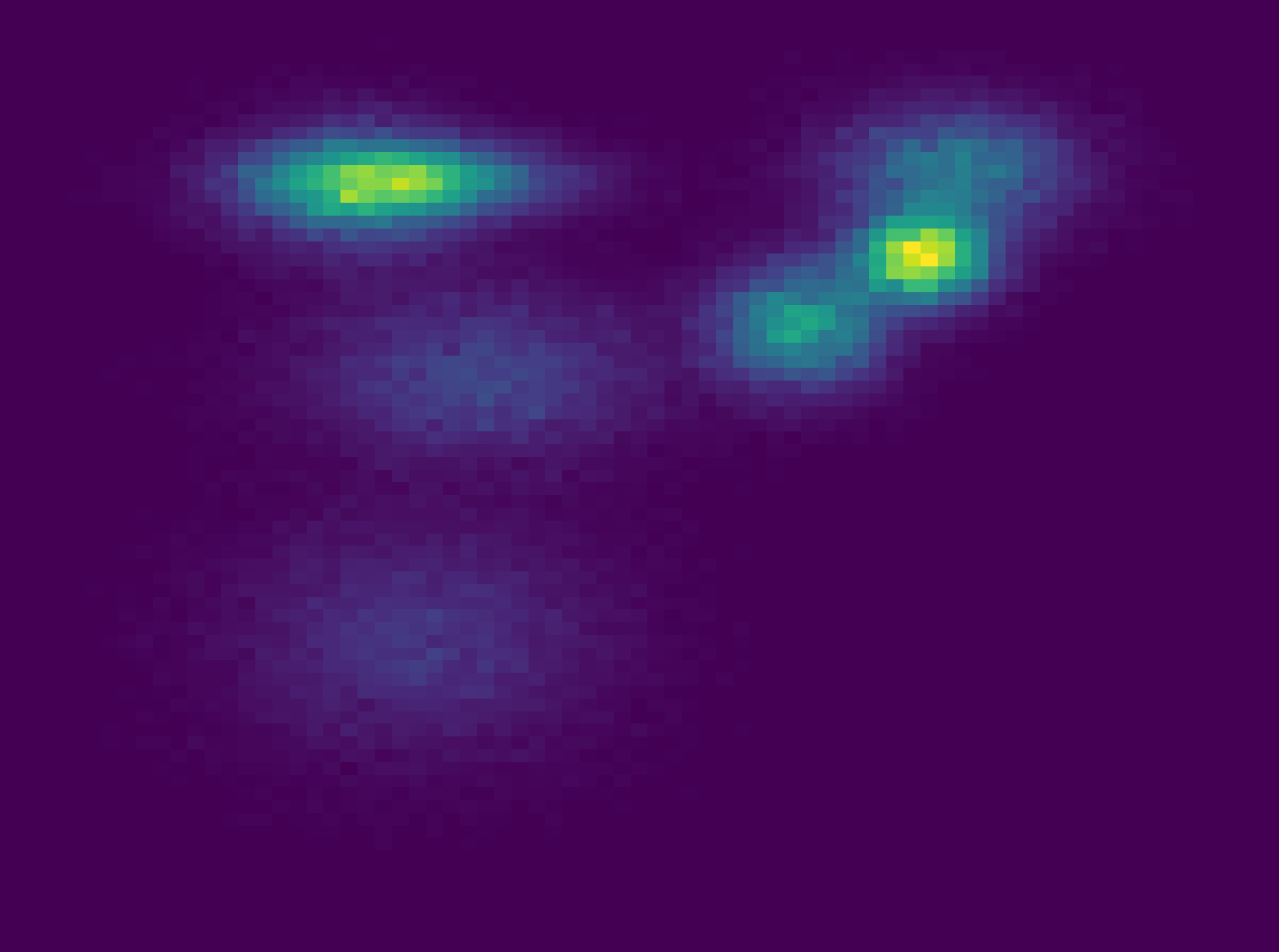}
    \caption{}
    \label{fig:dist-90k}
  \end{subfigure}
  \begin{subfigure}{0.24\textwidth}
    \includegraphics[width=\textwidth]{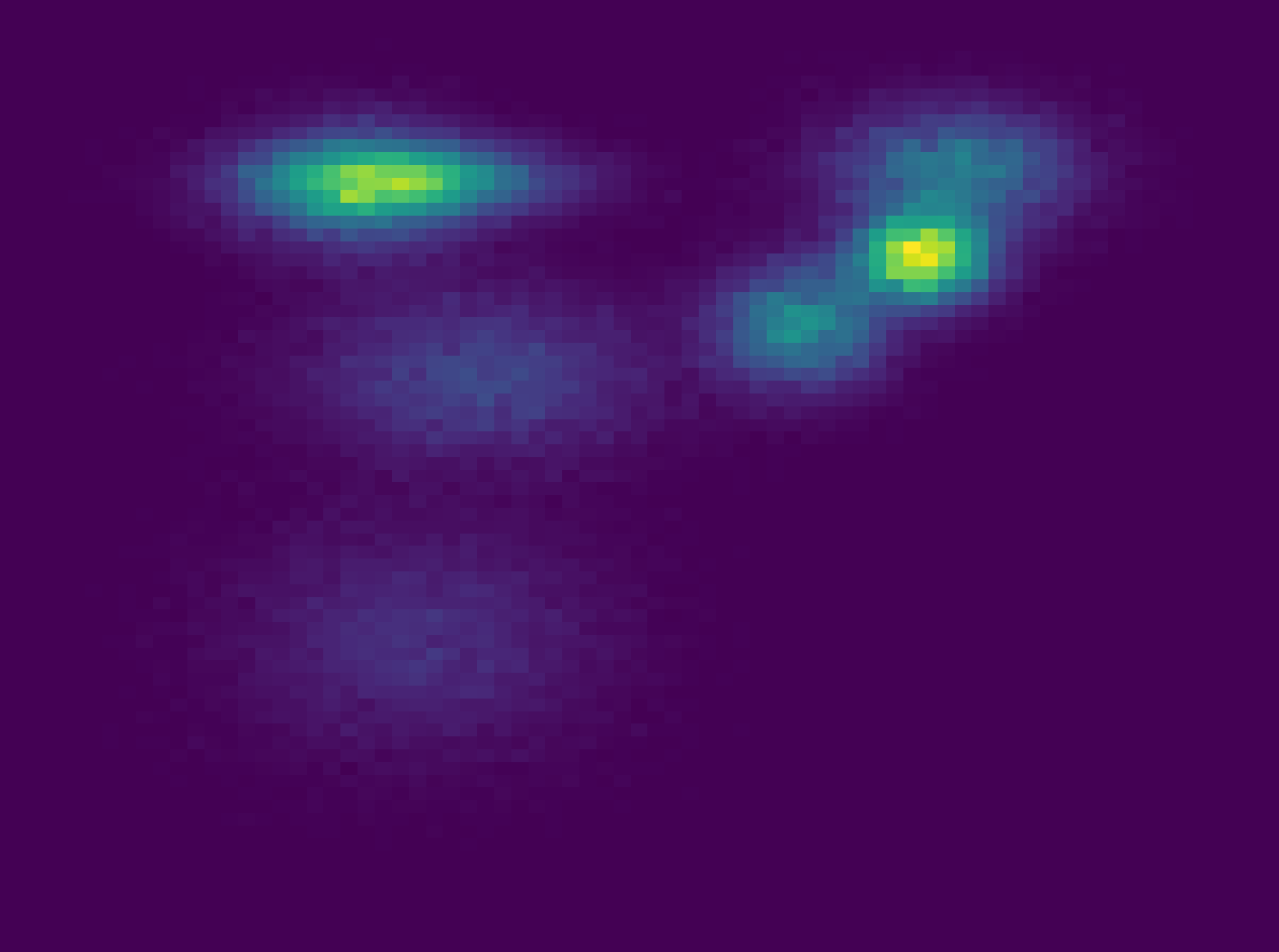}
    \caption{}
    \label{fig:dist-100k}
  \end{subfigure}
  \caption{
    Histograms of non-stationary simulated data. (\textbf{a}) shows
    the first 1000 observations, (\textbf{b}) shows the next 9000
    observations, (\textbf{c}) shows the last $90,000$
    observations, and (\textbf{d}) shows the total data set. In~each of
    the three time phases, the data-generating distribution produces data with
    more clusters than before.
  }
  \label{fig:simulated-dist-shift-data}
\end{figure}

As described in Section~\ref{sec:online}, changes in the data-generating
distribution should be detectable in the ML estimates. 
To investigate this, we generate simulated data with varying numbers of
clusters. 
Histograms of the simulated data can be seen in
Figure~\ref{fig:simulated-dist-shift-data}. 
The first 1000 observations were
generated from 3 Gaussian distributions, the next 9000 observations were
generated from 5 Gaussian distributions, including the 3 used to generate the
first observations, and the remaining 90,000 observations were generated from 7
Gaussian distributions, including the previous 5.
Some of the clusters overlap, so it is not immediately obvious from the
histograms how many clusters  there actually are.

We evaluate the effect of this type of distribution shift on SGAIS by estimating
the ML online for Gaussian mixture models with 3, 5, and 7 mixture components.
We then shuffle the data to enforce stationarity and estimate the ML for
these three models on the shuffled data.
If the final ML estimates for the in-order and shuffled data differ
significantly then this may indicate that the particles are getting trapped in
local modes before the change-points occur.

\section{Results and Discussion}

The log-ML typically grows linearly in the number of data points.
For this reason, it is natural to measure errors in $\log\ev/N$ rather
than $\log\ev$.
For each model, we measured the
runtime performance of NS, AIS, and SGAIS for various
data set sizes up to one million observations.
Each of these data sets is taken as the first $N$ observations of the largest
data set.
This allows SGAIS to leverage the previous ML estimates for online ML estimation.
Due to computational constraints, we run NS and AIS only for data set sizes at
logarithmically increasing intervals, while SGAIS naturally
produces many more intermediate results in a single run.
Since we found AIS to be much slower than NS, we only run AIS on data sets small
enough to finish within 4000 s.
We consider errors that are comparable to the discrepancy between NS and AIS
acceptable.

\subsection{Accuracy and Speed}
\unskip
\subsubsection{Linear Regression}
\begin{figure}[H]
  \begin{subfigure}{0.5\textwidth}
    \includegraphics[width=\textwidth]{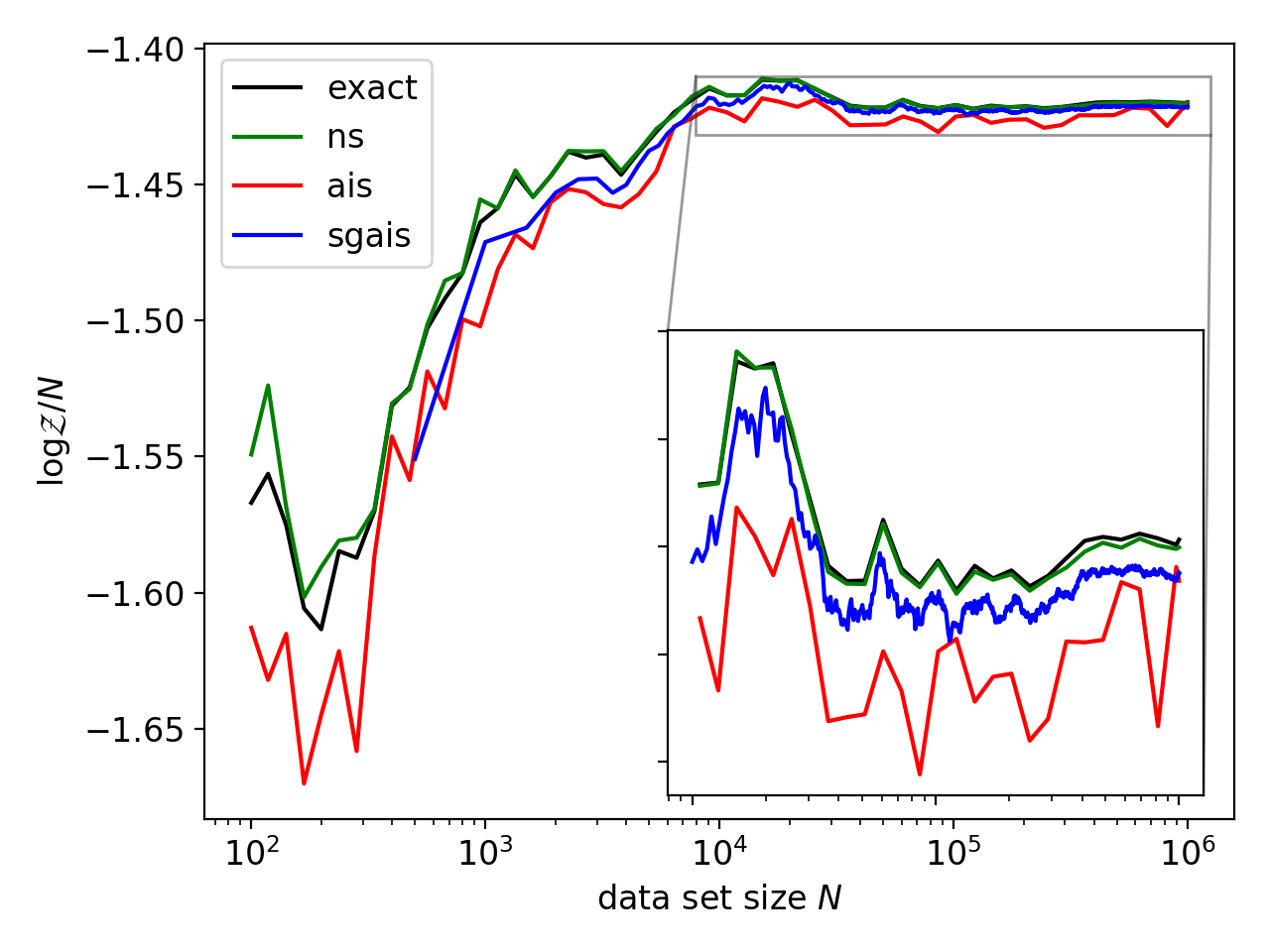}
    \caption{}
    \label{fig:lin-logz}
  \end{subfigure}
  \begin{subfigure}{0.5\textwidth}
    \includegraphics[width=\textwidth]{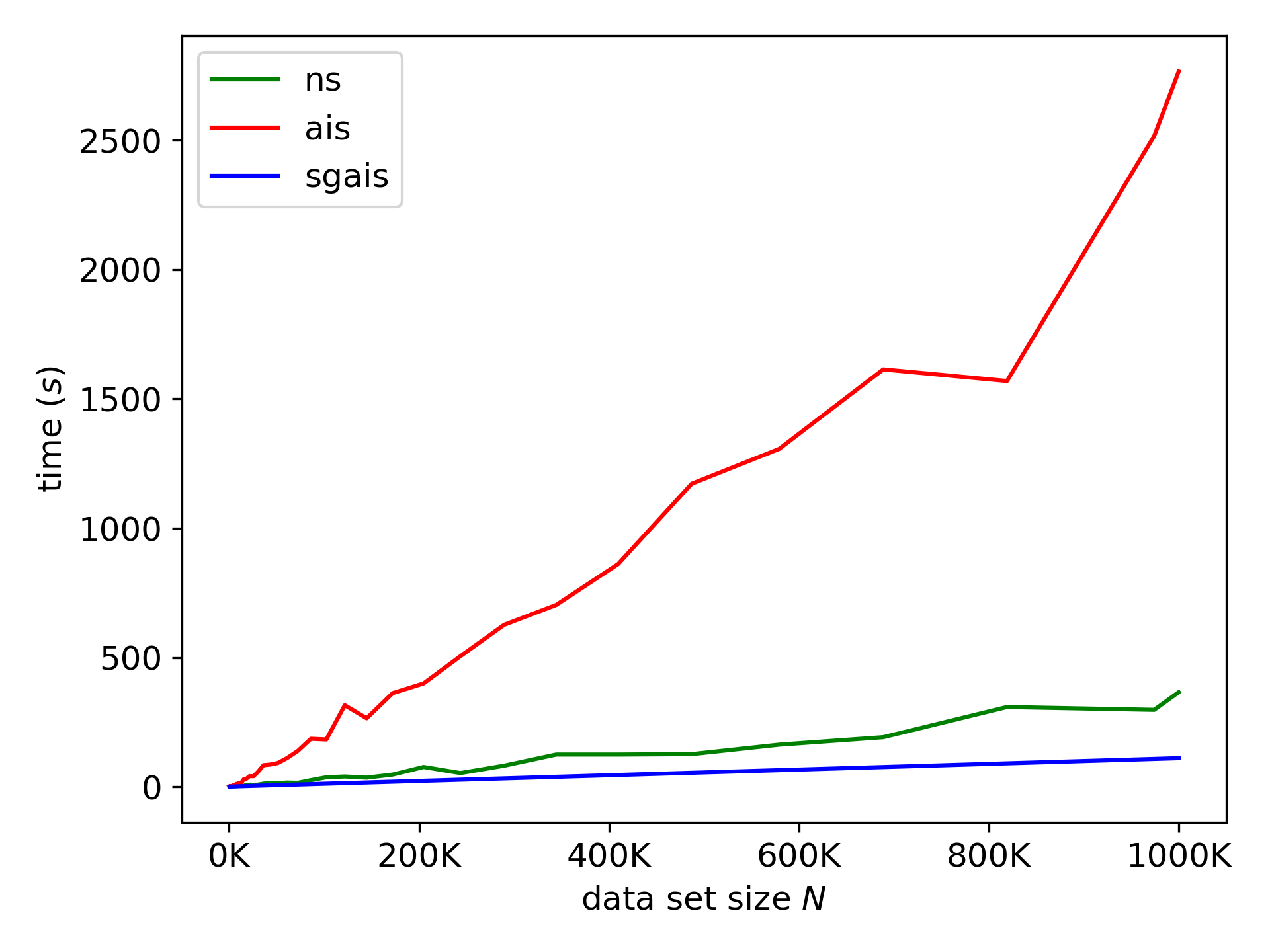}
    \caption{}
    \label{fig:lin-time}
  \end{subfigure}
  \caption{%
    Linear regression model. (\textbf{a}) shows the accuracy of our
    marginal likelihood (ML) estimator compared to nested sampling (NS), annealed importance sampling (AIS), and the exact ML.
    (\textbf{b}) shows the run-time of each method. \texttt{ns} is
    nested sampling, \texttt{ais} is annealed importance sampling, and
    \texttt{sgais} is our stochastic gradient annealed importance sampling
    approach.
  \label{f2}}
\end{figure}

For the linear regression model, the exact ML is available analytically and is
shown in Figure~\ref{f2}a for comparison. Each algorithm is able to
produce accurate results for this model for all data set sizes. 
The final error of SGAIS on one million data points was only
about 0.1\%.
For this model, our method achieved a speedup over NS by about a factor of 3.3,
and a speedup over AIS by a factor of 24.9 on one million observations.

\subsubsection{Logistic Regression}
\begin{figure}[H]
  \begin{subfigure}{0.5\textwidth}
    \includegraphics[width=\textwidth]{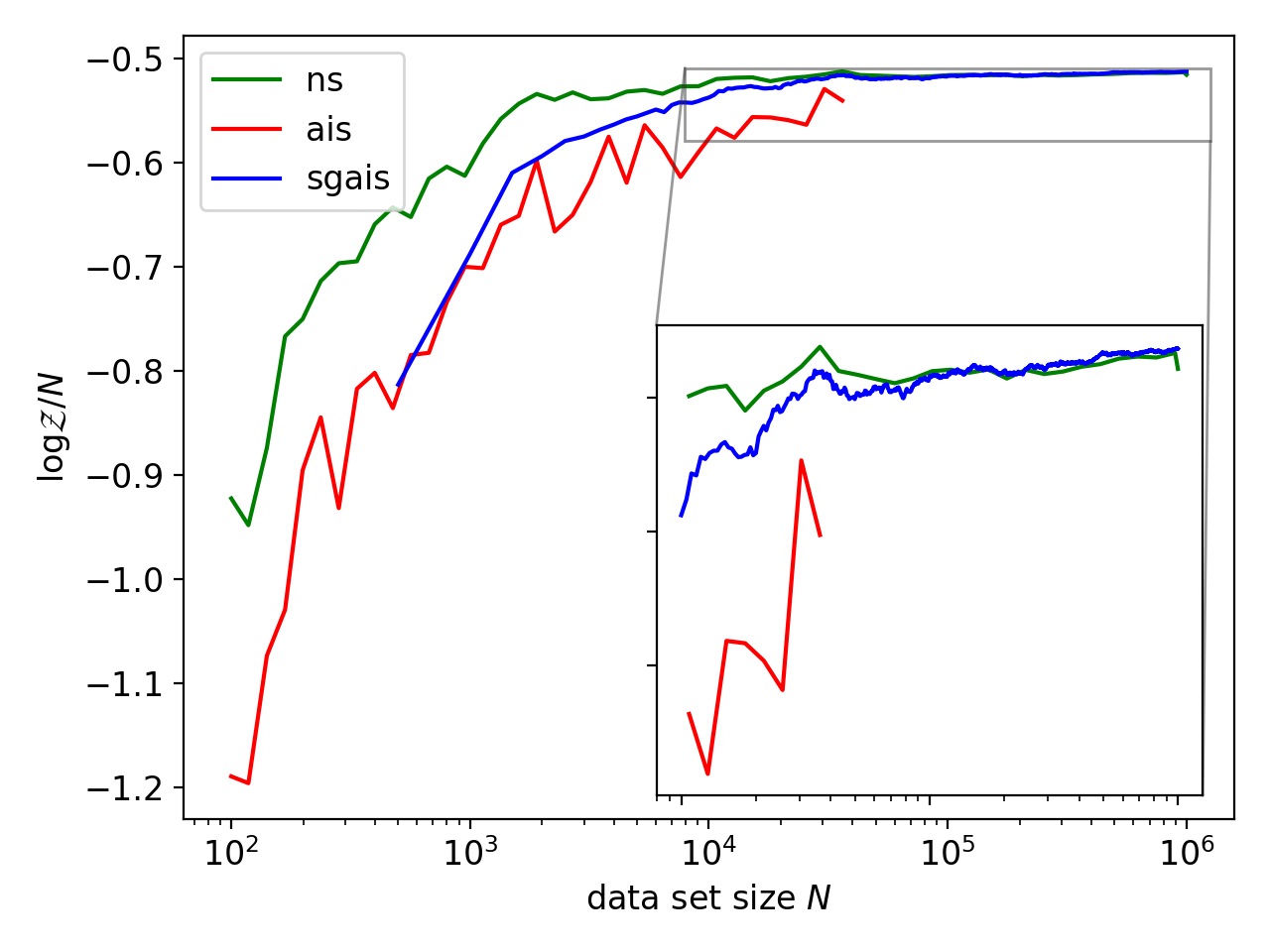}
    \caption{}
    \label{fig:log-logz}
  \end{subfigure}
  \begin{subfigure}{0.5\textwidth}
    \includegraphics[width=\textwidth]{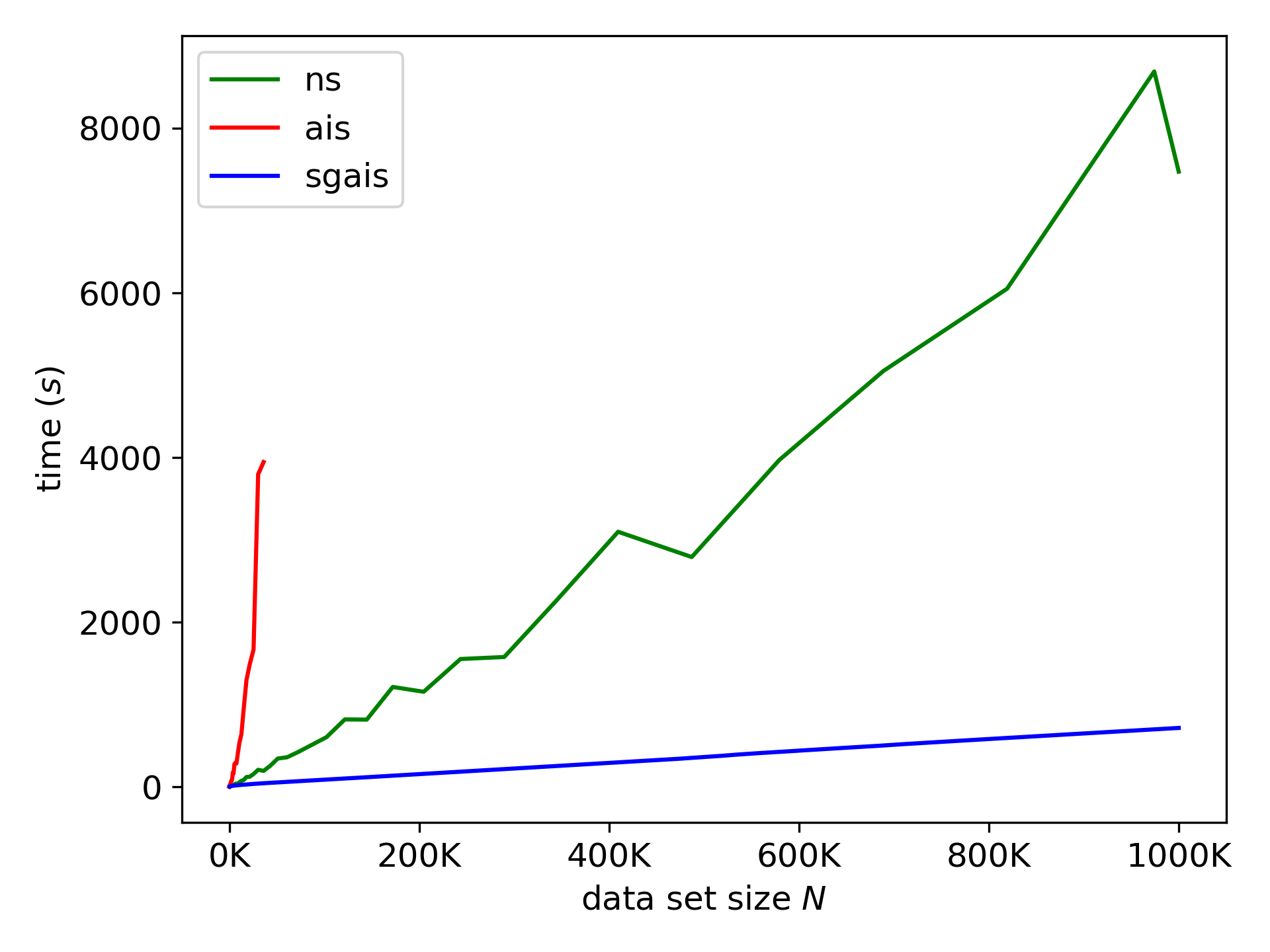}
    \caption{}
    \label{fig:log-time}
  \end{subfigure}
  \caption{%
    Logistic regression model. (\textbf{a}) shows our
    ML estimator compared to NS and AIS.
    (\textbf{b}) shows the run-time of each method. AIS was not run
    for larger data set sizes because each subsequent run would take more than
    4000 s.
  \label{f3}}
\end{figure}

Figure~\ref{f3}a,b shows the log-ML estimates and run-time of each algorithm for
the logistic regression model.
For the largest data set, NS and SGAIS produced estimates that
differed by roughly 0.6\%, which is negligible.
SGAIS was a factor 10.4 faster than the nested sampler on
one million observations for this model.

\subsubsection{Gaussian Mixture Model}
\begin{figure}[H]
  \begin{subfigure}{0.5\textwidth}
    \includegraphics[width=\textwidth]{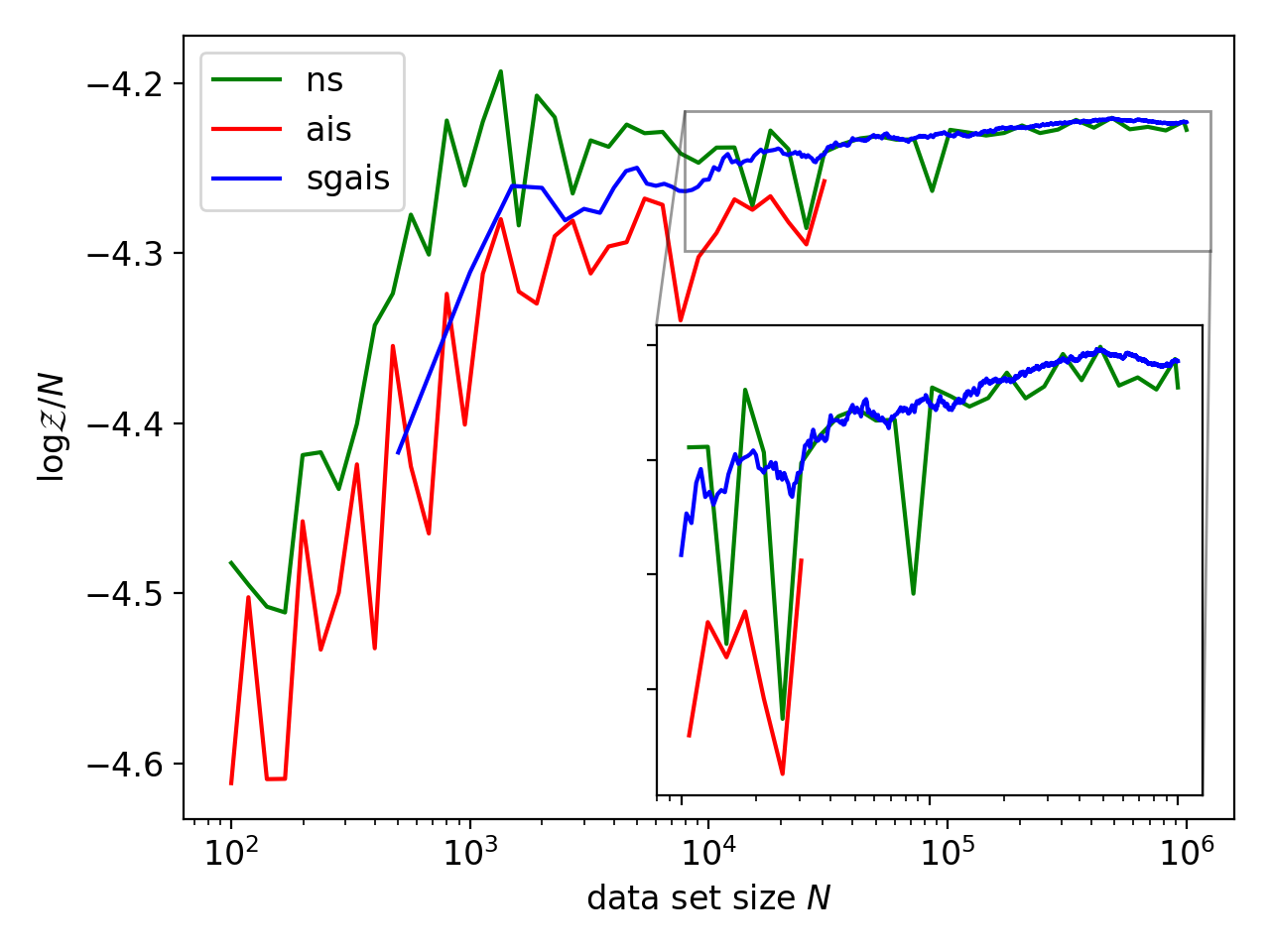}
    \caption{}
    \label{fig:gmm-logz}
  \end{subfigure}
  \begin{subfigure}{0.5\textwidth}
    \includegraphics[width=\textwidth]{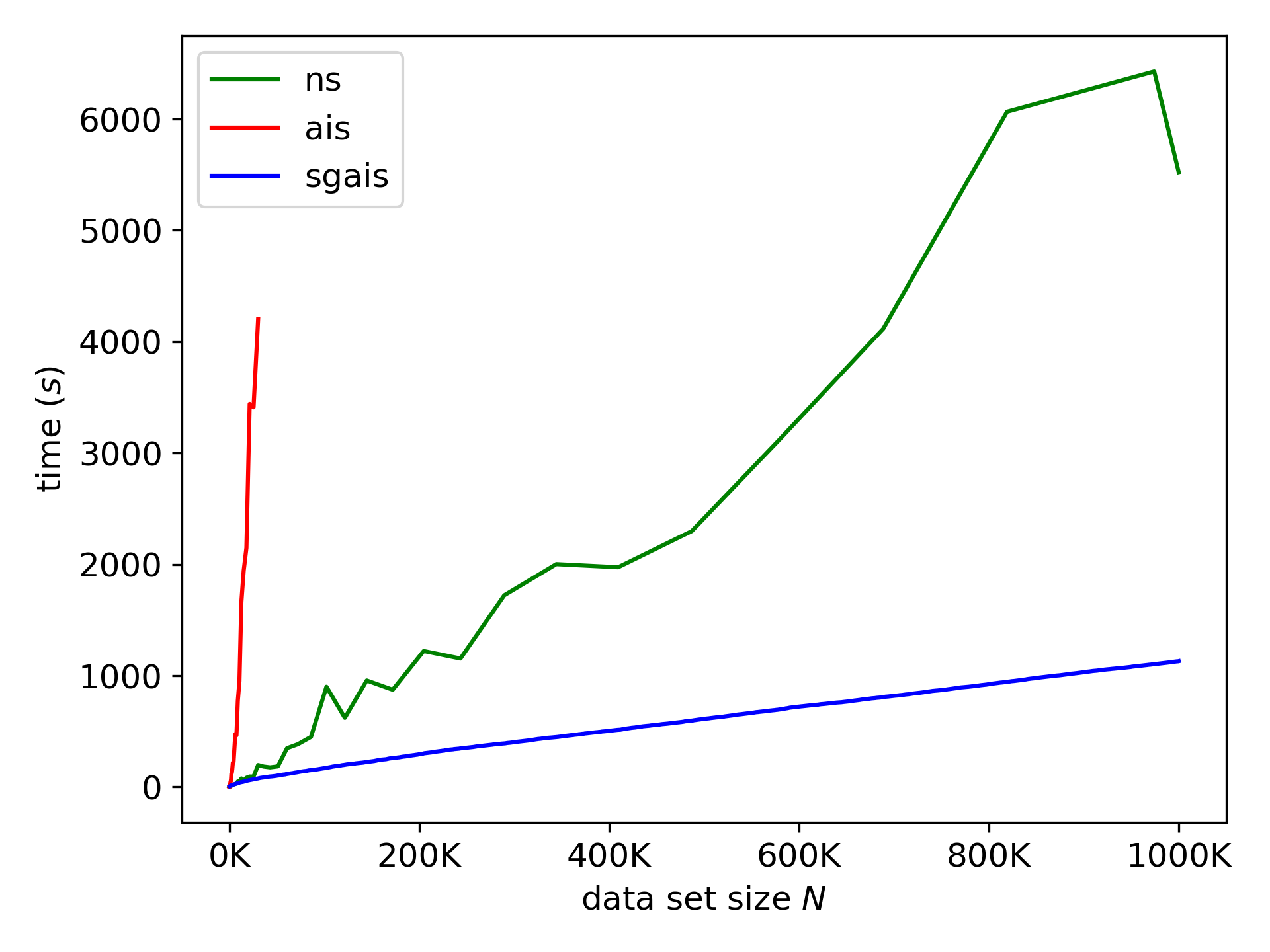}
    \caption{}
    \label{fig:gmm-time}
  \end{subfigure}
  \caption{%
    Gaussian mixture model. (\textbf{a}) shows our ML
    estimator compared to NS and AIS. (\textbf{b}) shows the
    run-time of each method.
    \label{f4}
  }
\end{figure}

Figure~\ref{f4}a,b shows the log-ML estimates and run-time of each algorithm for
the Gaussian mixture model.
The posterior distribution for this model is multimodal. Some modes are due to
permutation symmetries; these modes do not have to be  explored since each one
contains the same information. There are also some local modes that do not
necessarily capture meaningful information about the data; for example, fitting
a single Gaussian to the whole data set may be a poor local optimum of the
likelihood function. 
If an MCMC walker finds one of these modes, it can get trapped. However, we find
that by Bayesian updating and annealing, the MCMC walkers tend to leave the poor
local modes early on, before they become extremely peaked. 
The estimates produced by NS and SGAIS differed on the largest data set by
roughly 0.1\%. For this model, SGAIS was about a factor of 4.9
faster than the nested sampler for one million observations.

In all the above experiments, SGAIS converges to the same
result as NS with a negligible error for large $N$.

\subsection{Distribution Shift}

\begin{figure}[H]
  \begin{subfigure}{0.5\textwidth}
    \includegraphics[width=\textwidth]{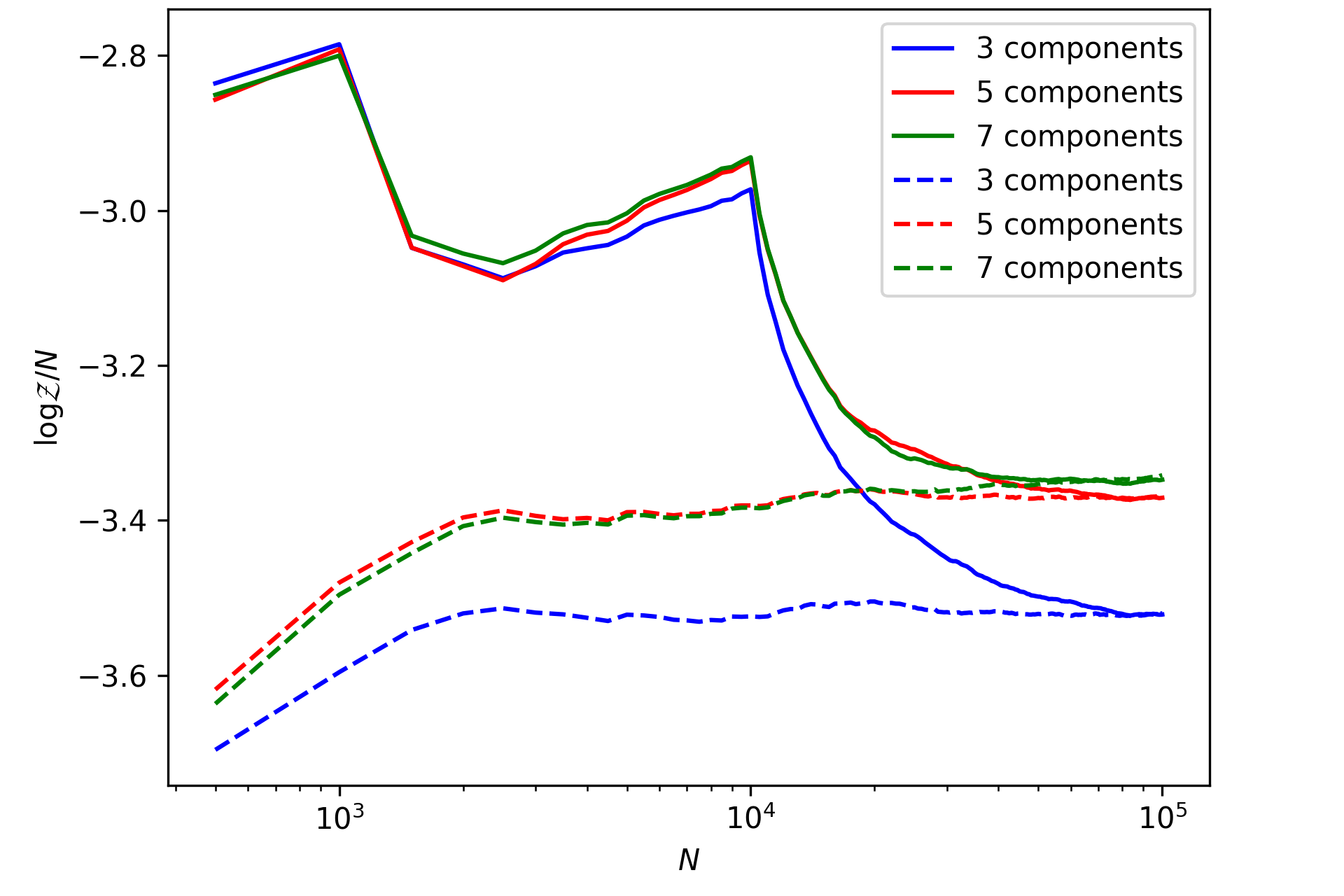}
    \caption{}
    \label{fig:dist-shift-logz}
  \end{subfigure}
  \begin{subfigure}{0.5\textwidth}
    \includegraphics[width=\textwidth]{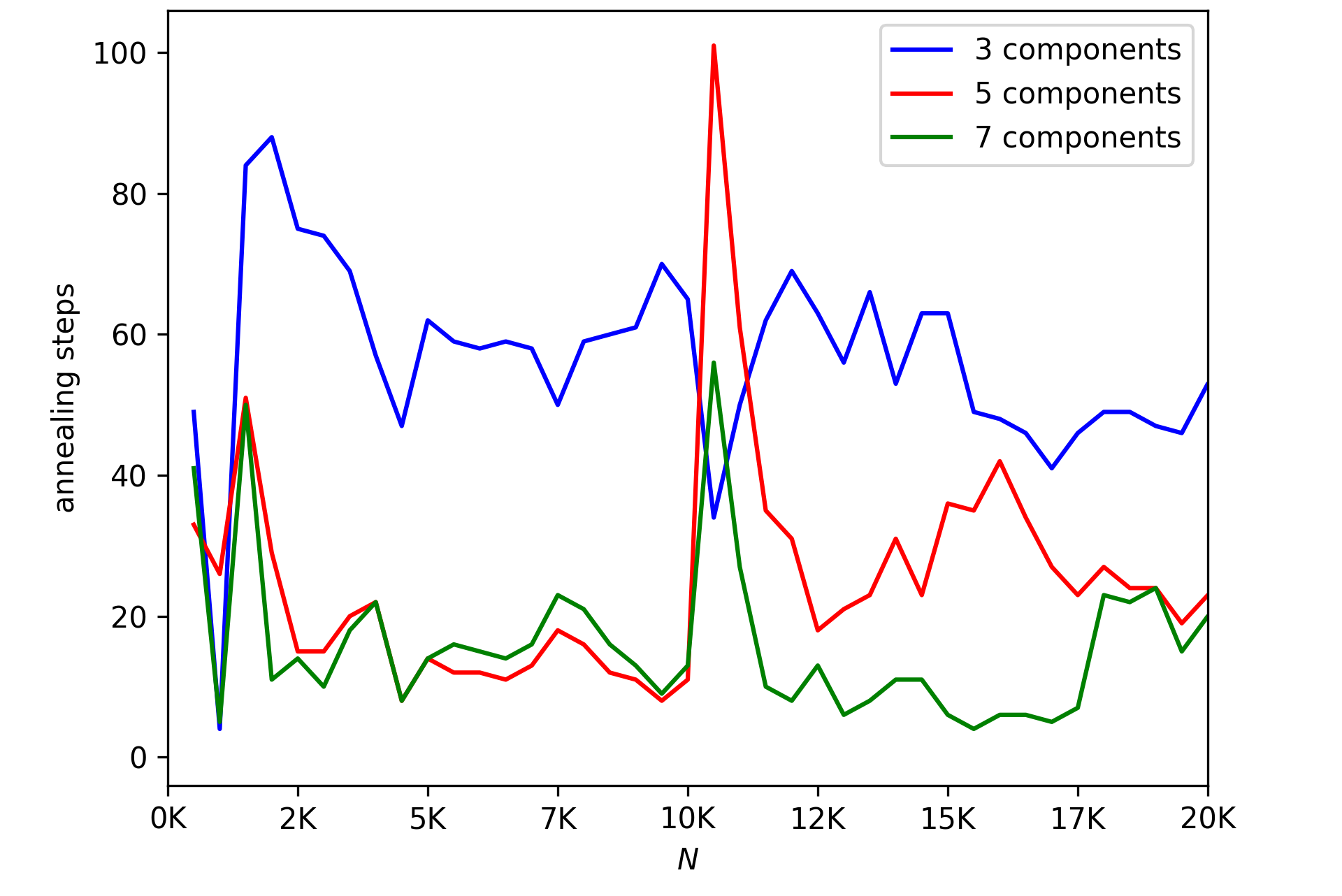}
    \caption{}
    \label{fig:dist-shift-steps}
  \end{subfigure}
  \caption{
    ML estimation under distribution shift.
    (\textbf{a}) shows the ML estimates for Gaussian
    mixture models with different numbers of mixture components.
    The solid lines are for the in-order data and the dashed lines are for the
    shuffled
    and therefore stationary data.
    (\textbf{b}) shows the number of annealing steps for the
    online ML estimates for the in-order data.
  \label{f5}}
\end{figure}

The log-ML estimates shown in Figure~\ref{f5}a display sharp
changes at 1000 and 10,000 observations for the non-shuffled data.
We are very clearly able to identify the position of the change-points just by
looking at the resulting plot, without a priori assuming the existence or number of
change-points.
The~numbers of annealing steps shown in Figure~\ref{f5}b exhibit
spikes at the change-points and remain high once more clusters are added to the
data than the model can describe.

The agreement of the final ML estimates between the shuffled and non-shuffled
data suggests that these estimates can be trusted.
The difference between the online and shuffled estimates is small enough to
be able to distinguish between the three models.
The 5- and 7-component models seem to describe the total data set better than the
3-component model, but the 5 and 7 models have similar values for their log-ML,
presumably due to the overlapping clusters in the data set.

\subsection{Sensitivity to Algorithm Parameters}

The following results agreed closely with our expectations:
\begin{itemize}[leftmargin=*,labelsep=5.8mm]
  \item Increasing the number of particles, $M$, results in higher accuracy and a
    longer running time without much effect on the number of annealing steps. 
  \item A smaller number of burn-in SGHMC steps per intermediate distribution
    typically resulted in lower accuracy and a shorter run-time. A smaller number
    of burn-in steps also resulted in more annealing steps due to the slower
    equilibration.
  \item Larger mini-batch sizes typically result in higher accuracy but more
    computation per SGHMC step. Larger mini-batch sizes result in fewer Bayesian
    updating steps but require more annealing steps per new chunk of data.
    Mini-batch size would typically be chosen based on the hardware capabilities
    of the platform and the type of data under consideration.
\end{itemize}

We report on our more interesting findings below; plots of the results for each
parameter investigated are given in Appendix~\ref{ap:figs}.

\subsubsection{Target $\ESS$}
\begin{figure}[H]
  \begin{subfigure}{0.33\textwidth}
    \includegraphics[width=\textwidth]{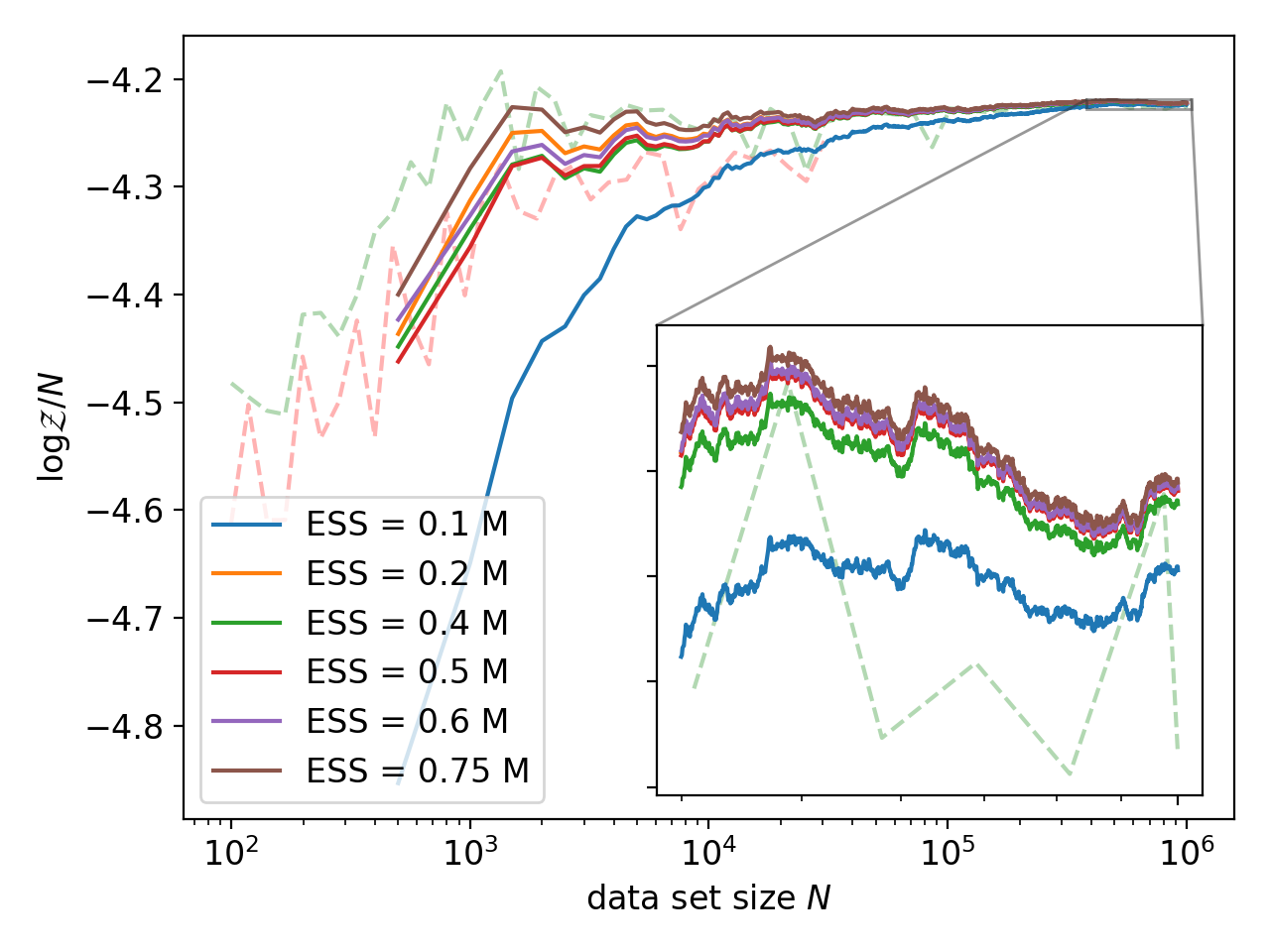}
    \caption{}
    \label{fig:ess-logz}
  \end{subfigure}
  \begin{subfigure}{0.33\textwidth}
    \includegraphics[width=\textwidth]{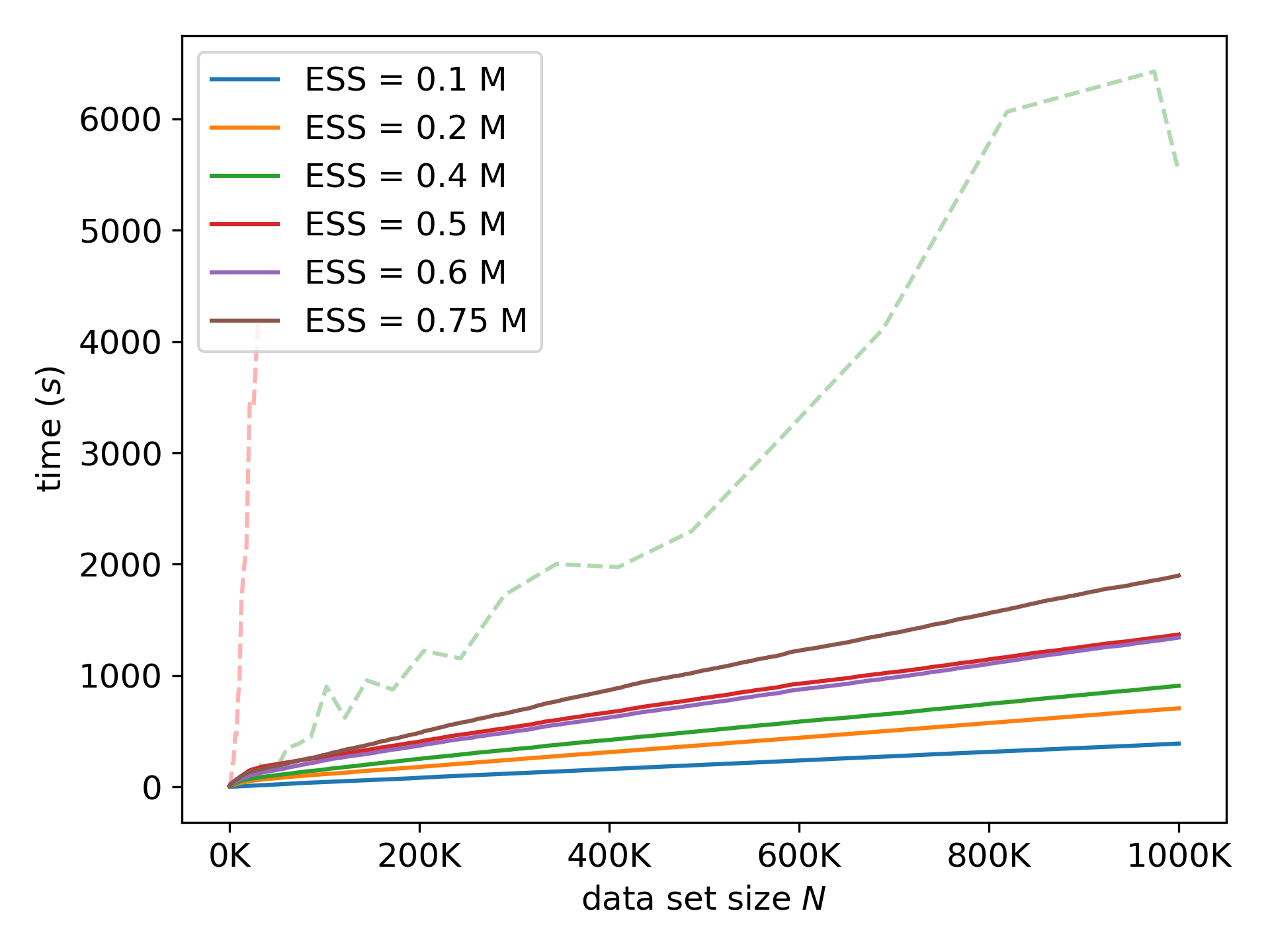}
    \caption{}
    \label{fig:ess-time}
  \end{subfigure}
  \begin{subfigure}{0.33\textwidth}
    \centering
    \includegraphics[width=\textwidth]{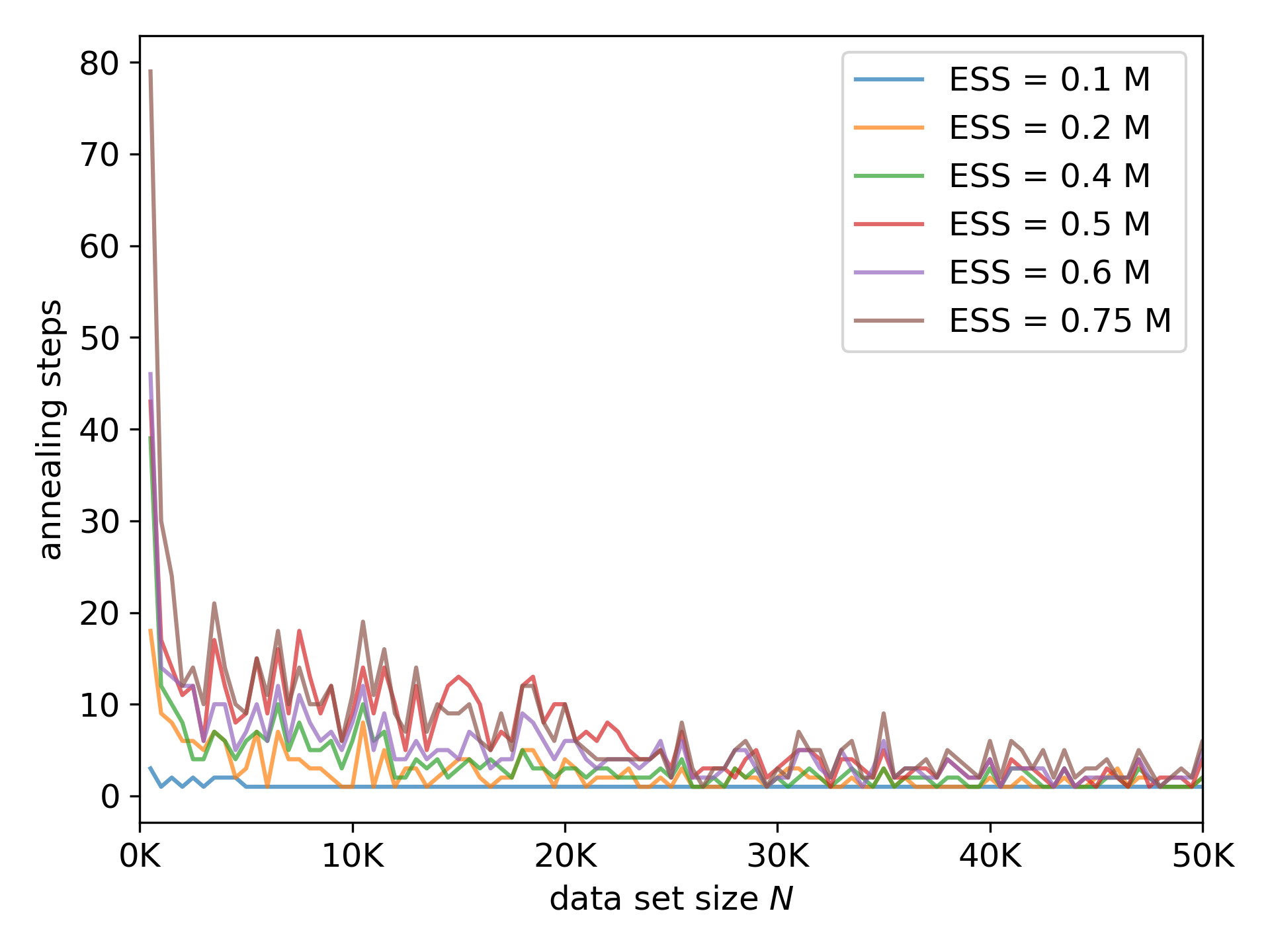}
    \caption{}
    \label{fig:ess-steps}
  \end{subfigure}
  \caption{%
    Sensitivity to the target effective
    sample size ($\ESS$).
    (\textbf{a}) shows the log-ML estimates, (\textbf{b})
    shows the run-time, and (\textbf{c}) shows the number of annealing
    steps for each chunk of data against the data set size until that chunk.
  \label{f6}}
\end{figure}

As expected, a higher target $\ESS$ tends to result in more annealing steps---see Figure~\ref{f6}a.
Most of this work is done in the early stages of Bayesian updating.
Note that since we used 10 particles, a target $\ESS$ of $0.1 M = 1$
requires no annealing steps because $\ESS$ is bounded below by 1.
%
%
No annealing results in high variance during the early stages of Bayesian
updating, and adaptively annealing helps to reduce that variance, with only a
small impact on the run-time.
This~illustrates the importance of the adaptive annealing schedule in our
approach.
Figure~\ref{f6}a,b indicates that our approach
converges to the log-ML within acceptable accuracy within a reasonable time for
a target $\ESS$ larger than 1.
Even a small target $\ESS$ was good enough to match vanilla AIS, on average.

\subsubsection{Learning Rate}
\begin{figure}[H]
  \begin{subfigure}{0.33\textwidth}
    \includegraphics[width=\textwidth]{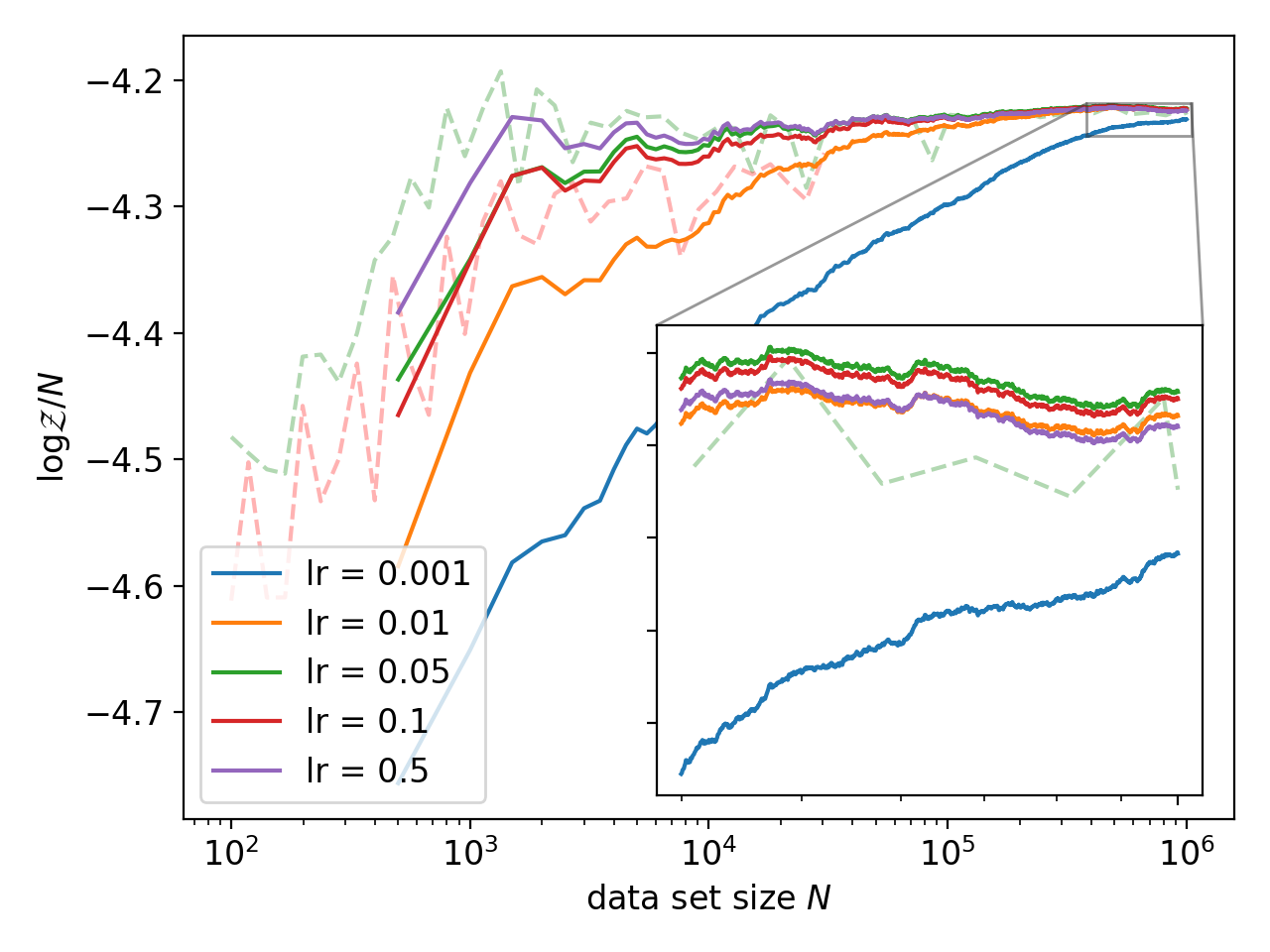}
    \caption{}
    \label{fig:lr-logz}
  \end{subfigure}
  \begin{subfigure}{0.33\textwidth}
    \includegraphics[width=\textwidth]{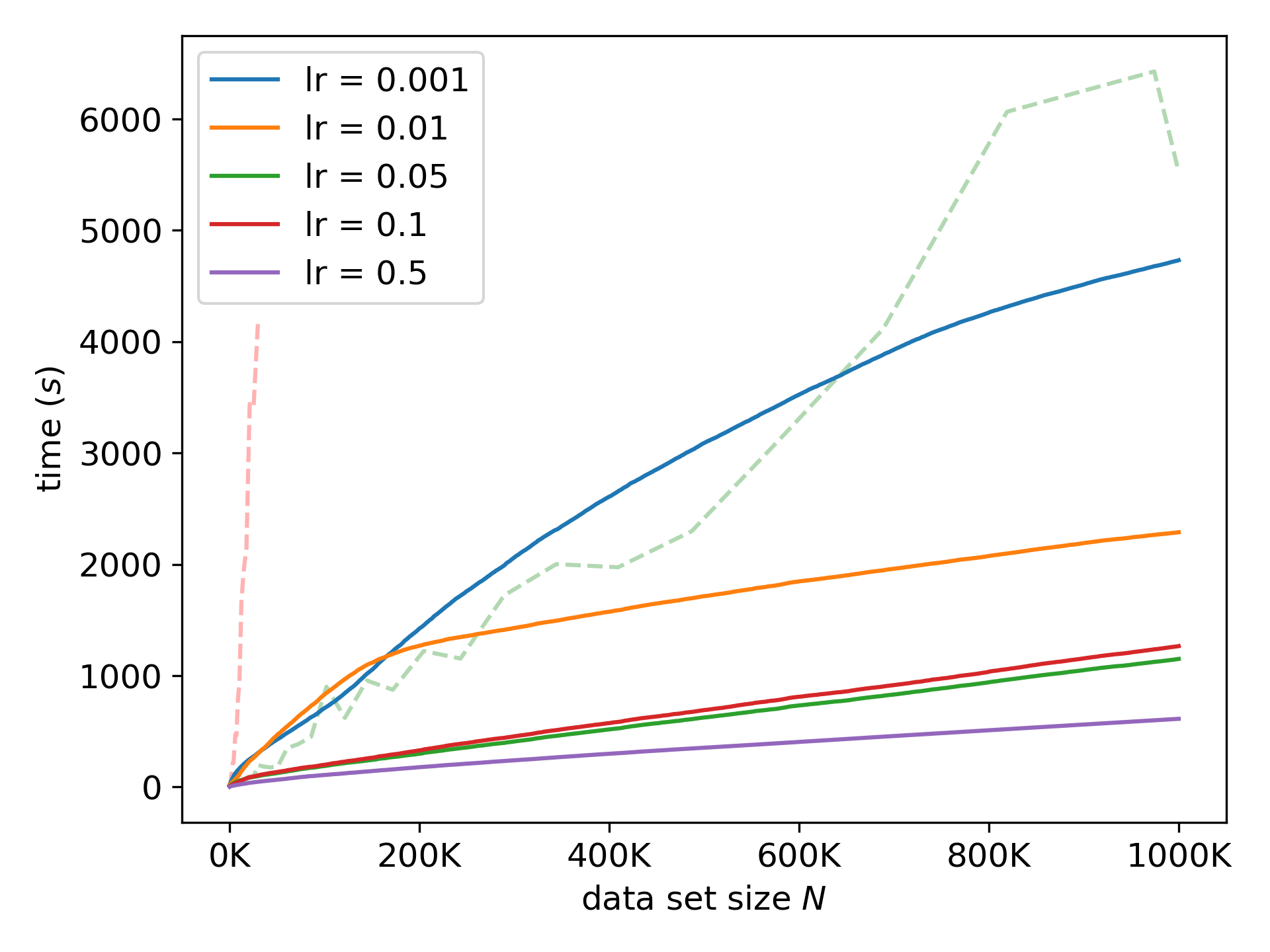}
    \caption{}
    \label{fig:lr-time}
  \end{subfigure}
  \begin{subfigure}{0.33\textwidth}
    \centering
    \includegraphics[width=\textwidth]{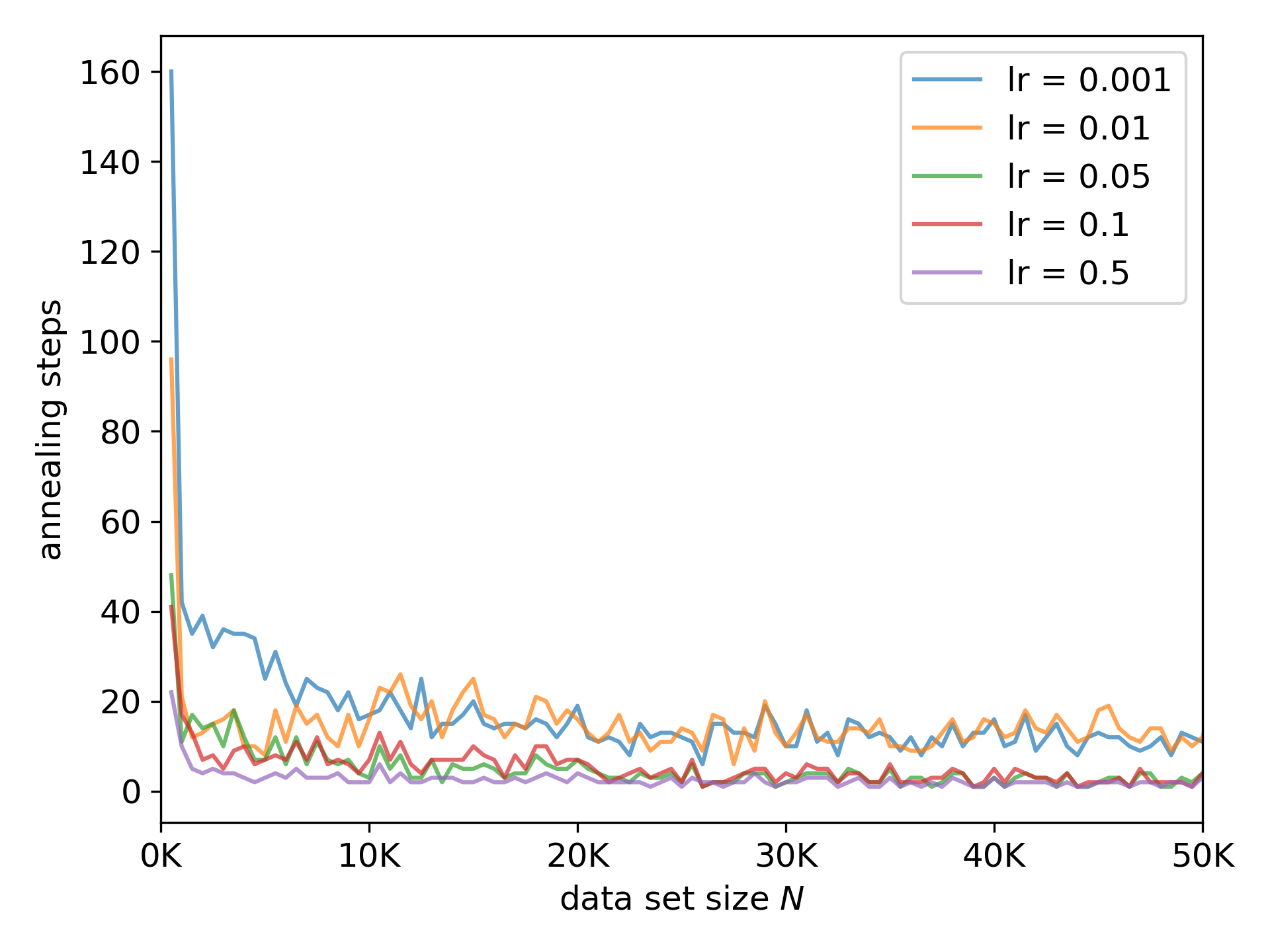}
    \caption{}
    \label{fig:lr-steps}
  \end{subfigure}
  \caption{%
    Sensitivity to the learning rate. Reported learning rates are
    per-observation learning rates, that is, $\eta = \mathrm{lr}/N$.
    (\textbf{a}) shows the log-ML estimates, (\textbf{b})
    shows the run-time, and (\textbf{c}) shows the number of annealing
    steps. For a learning rate of 0.01, the run-time shown in
    (\textbf{b}) displays a change in gradient near $10^5$
    observations. This is the result of a reduced number of annealing steps but
    is not visible in (\textbf{c}) since we only show the number of
    annealing steps for up to $5\times10^4$ observations.
  \label{f7}}
\end{figure}

Interestingly, smaller values of the learning rate tend to result in less
accurate log-ML estimates over a longer time---see
Figure~\ref{f7}a,b.
We suspect this to be because a smaller learning rate does not allow the particle
to move as far each step, resulting in a slower equilibration and requiring more
annealing steps per observation.
This effect can be seen in Figure~\ref{f7}c; the smaller learning rates
appear to result in a larger number of annealing steps per
observation.
To further verify this, we investigated the interaction between the learning
rate and the number of burn-in steps.

\subsubsection{Learning Rate and Burn-in}
\begin{figure}[H]
  \begin{subfigure}{0.33\textwidth}
    \includegraphics[width=\textwidth]{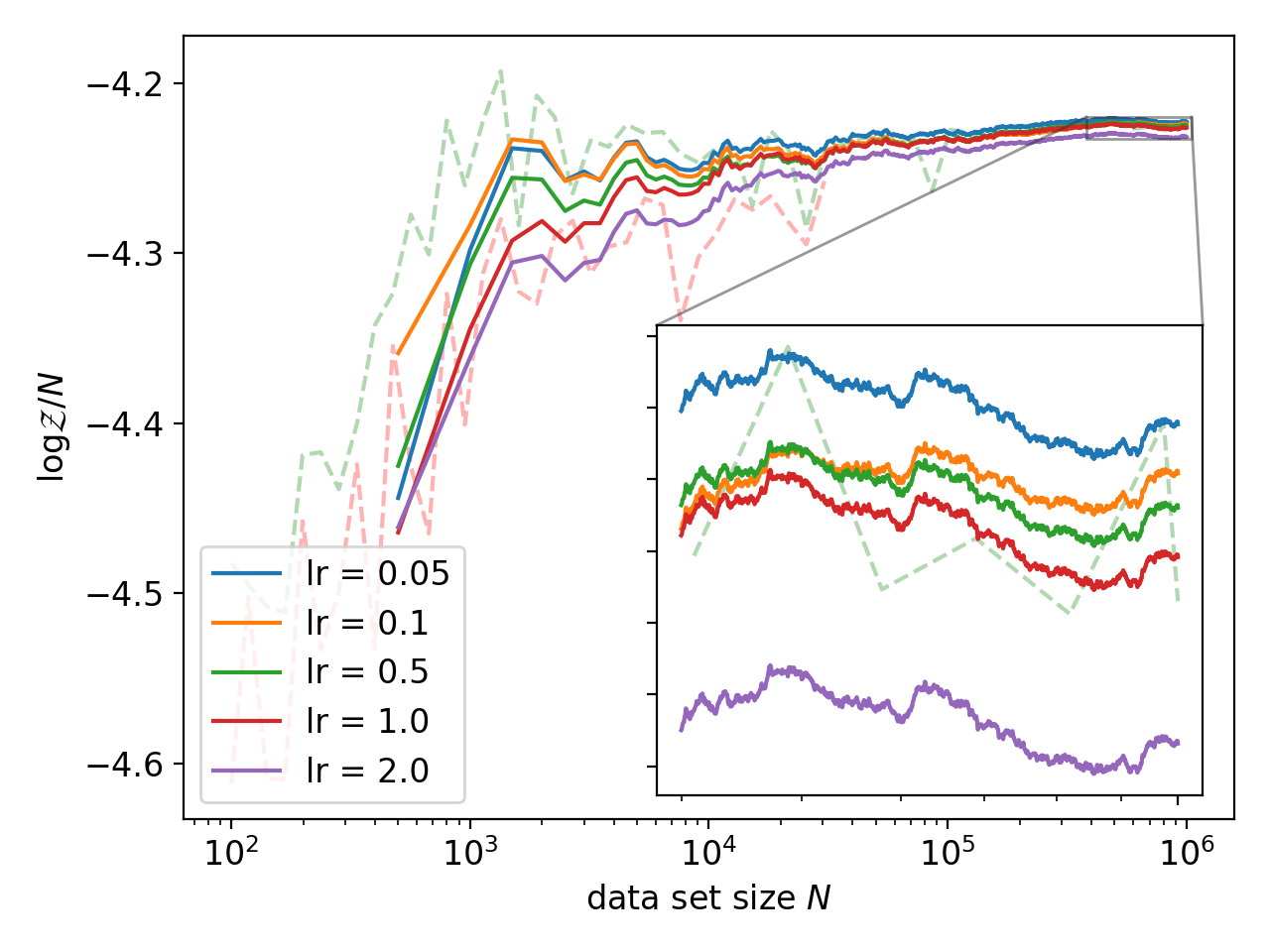}
    \caption{}
    \label{fig:lr-burnin-logz}
  \end{subfigure}
  \begin{subfigure}{0.33\textwidth}
    \includegraphics[width=\textwidth]{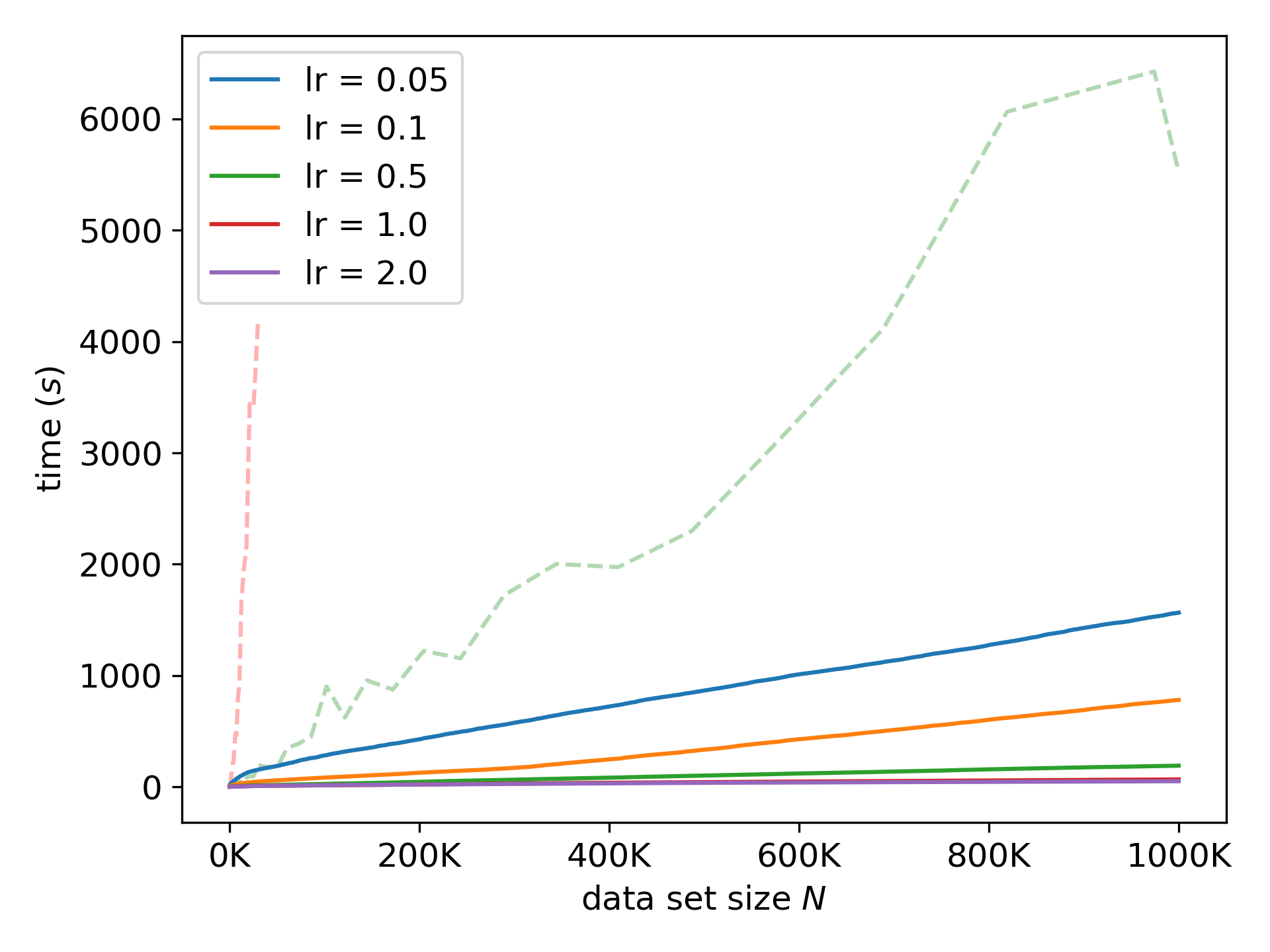}
    \caption{}
    \label{fig:lr-burnin-time}
  \end{subfigure}
  \begin{subfigure}{0.33\textwidth}
    \centering
    \includegraphics[width=\textwidth]{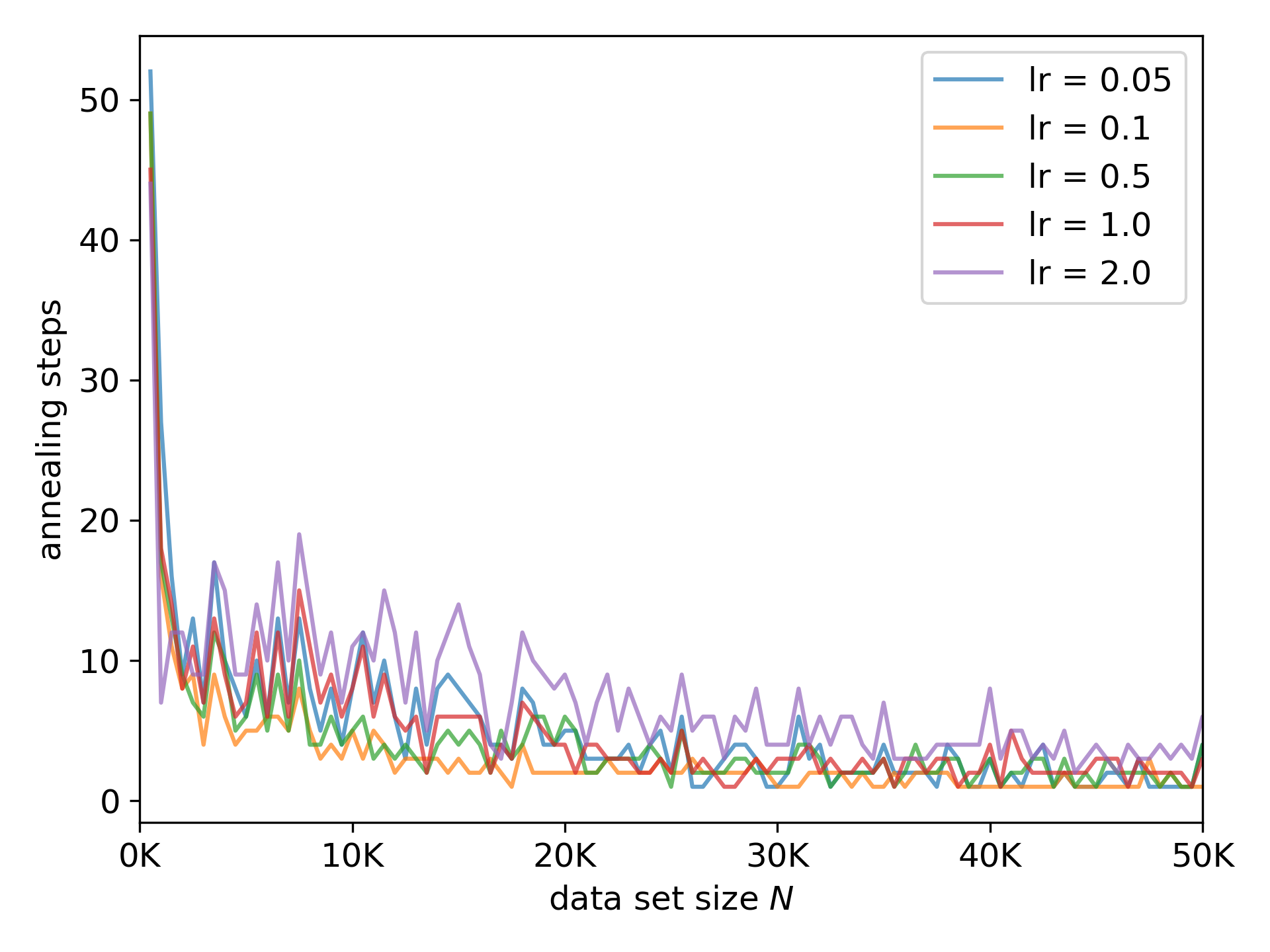}
    \caption{}
    \label{fig:lr-burnin-steps}
  \end{subfigure}
  \caption{%
    Sensitivity to the learning rate while keeping the product of the learning
    rate and the number of stochastic gradient Hamiltonian Monte
    Carlo (SGHMC) steps constant. Reported learning rates are
    per-observation learning rates, that is, $\eta = \mathrm{lr}/N$.
    (\textbf{a}) shows the log-ML estimates,
    (\textbf{b}) shows the run time, and
    (\textbf{c}) shows the number of annealing steps.
  \label{f8}}
\end{figure}

We investigate the interaction between the number of SGHMC steps taken per
intermediate distribution and the learning rate by
varying the learning rate, while keeping the product of the
learning rate and the number of SGHMC steps constant.
Fewer burn-in steps (larger learning rate) tends to make the
algorithm faster, but a smaller learning rate results in higher accuracy in the
log-ML estimates, as seen in Figure~\ref{f8}a.
The decrease in accuracy with a larger learning rate is presumably due to the
discretization error in Equation~(\ref{eq:sghmc}).
This is supported by 
Figure~\ref{f8}c: a larger learning rate requires a larger
number of annealing steps to reach the target $\ESS$.
Figure~\ref{f8}a indicates that a per-observation learning rate
of 1.0 can be used and still result in estimates of acceptable accuracy on data
sets of one million observations. 
For a learning rate of 1.0, SGAIS achieved a speedup over NS by a factor of 83.

\section{Materials and Methods}

Code for this work was implemented using
pytorch~\cite{pytorch}. The code for our experiments is
available at \url{https://gitlab.com/pleased/sequential-evidence}.

\section{Conclusions}

This paper introduced SGAIS, a novel algorithm for efficient large-scale ML
estimation, by combining the essential ingredients of Bayesian updating,
thermal annealing and data sub-sampling.%
We found that our SGAIS implementation was able to
produce accurate ML estimates with a speedup over NS and AIS on simple models.
Furthermore, since the
marginal cost of updating the ML estimates when new data arrives does not depend
on the number of previous data points, this approach may be effective for
weighted model averaging in a setting where one periodically gets access to new
data, such as in streaming applications.

We evaluated the sensitivity of our approach to the parameters and found that the
log-ML estimates were robust to many different parameter choices while
maintaining a speedup over NS and AIS for large data sets.

Potential future work may include further exploring the
effects of stochastic gradients for ML calculation in the full SMC setting,
including latent variable models, and with more general
stochastic gradient MCMC algorithms such as Riemannian manifold
SGHMC~\cite{sgrhmc}.

Lastly, we would like to point out that SGAIS naturally inherits some of the
pitfalls of AIS and SGHMC such as sensitivity to phase changes, local modes,
and the curse of dimensionality.
Bayesian updating and adaptive scheduling does help to reduce some of the
challenges that come with annealing, just as annealing helps to reduce problems
arising from multi-modality, but to our knowledge, there is no way to
completely remove these difficulties in general.

\vspace{6pt} 


\authorcontributions{
%
Conceptualization,
methodology,
software,
formal analysis,
investigation,
visualization,
writing---original draft preparation: S.A.C.;
validation,
writing---review and editing: S.A.C., H.C.E., and S.K.;
resources: S.K.;
project administration,
supervision: H.C.E. and S.K.;
funding acquisition: H.C.E.
%
}

\funding{S. Cameron received bursary support from the South African National
Institute of Theoretical Physics.}

\acknowledgments{We wish to thank the South
African National Institute of Theoretical Physics for financial support and the
organizers of MaxEnt~2019 for sponsoring the publication fee for this paper.}

\conflictsofinterest{The authors declare no conflict of interest.}

\abbreviations{The following abbreviations are used in this manuscript:\\

\noindent 
\begin{tabular}{@{}ll}
AIS & Annealed importance sampling\\
ESS & Effective sample size\\
i.i.d & Independently and identically distributed\\
MCMC & Markov chain Monte Carlo\\
ML & Marginal likelihood\\
NS & Nested sampling\\
SGAIS & Stochastic gradient annealed importance sampling\\
SGHMC & Stochastic gradient Hamiltonian Monte Carlo\\
SMC & Sequential Monte Carlo
\end{tabular}}

\appendixtitles{yes} 
\appendix
\section{Correctness of Nested Sampler and Annealed Importance Sampler\label{ap:ns}}
\unskip
\ifdraft{}{
\subsection{Nested Sampler Implementation}

Nested sampling requires one to sample from the prior under increasing
likelihood constraints. Sampling under constraints is, in general, a difficult
problem. This sampling can be accomplished by repeatedly sampling from the prior
until the constraint is satisfied. This method of rejection sampling scales
extremely poorly with both the size of the data set and the dimension of the
parameter space and is not feasible on any but the smallest problems.
Instead, the constrained sampling is typically implemented by duplicating a live
particle, which is already within the constrained likelihood contour, and
applying an MCMC kernel to the particle that leaves the prior invariant.
While moving the particle with MCMC, care must be taken to avoid leaving the
constrained region.
\newpage
Our implementation of NS is based on SGHMC, simply because the code for SGHMC
was already written. In order to sample under constraints, we use a similar
strategy to Galilean Monte Carlo as described in~\cite{gmc-ns}, where the
particle reflects off of the boundary of the likelihood contour by updating the
momentum as follows:
\begin{equation*}
  \Delta\ubv = -2 (\ubv\cdot\ubn) \ubn.
\end{equation*}
Here, $\ubn$ is a unit vector parallel to the likelihood gradient
$\nabla_{\bmtheta} p(\data|\bmtheta)$. While it is not the focus of our paper,
we note that we have not previously encountered the idea of applying SGHMC to
sampling under constraints; our implementation allows a fairly direct approach
to implementing NS in a variety of contexts, and the technique may also
potentially be of value in other constrained sampling contexts.

We found that due to discretization error, the constrained sampler tends to
undersample slightly at the boundaries of the constrained region; however, the
undersampled volume is of the order of the learning rate $\eta$ and so can be
neglected for a sufficiently small learning rate.

\subsection{Test Results}

If either of our NS and AIS implementations were incorrect, we expect that their
ML estimates would disagree. 
If both implementations were incorrect, the probability that their errors would
cancel is close to zero since NS and AIS work in completely different ways.
Figure~\ref{fig:ns-ais} shows the result of 5~independent runs of NS and AIS for
data sets of varying sizes from 20 to 1000, in intervals of 20.
The~shaded regions are the minimum and maximum log-ML estimates for each run;
these regions are the typical values produced by NS and AIS on these data sets.
Since the typical results of NS and AIS tend to agree, it is safe to assume that
they have been implemented correctly.

\begin{figure}[H]
  \centering
  \includegraphics[width=0.7\textwidth]{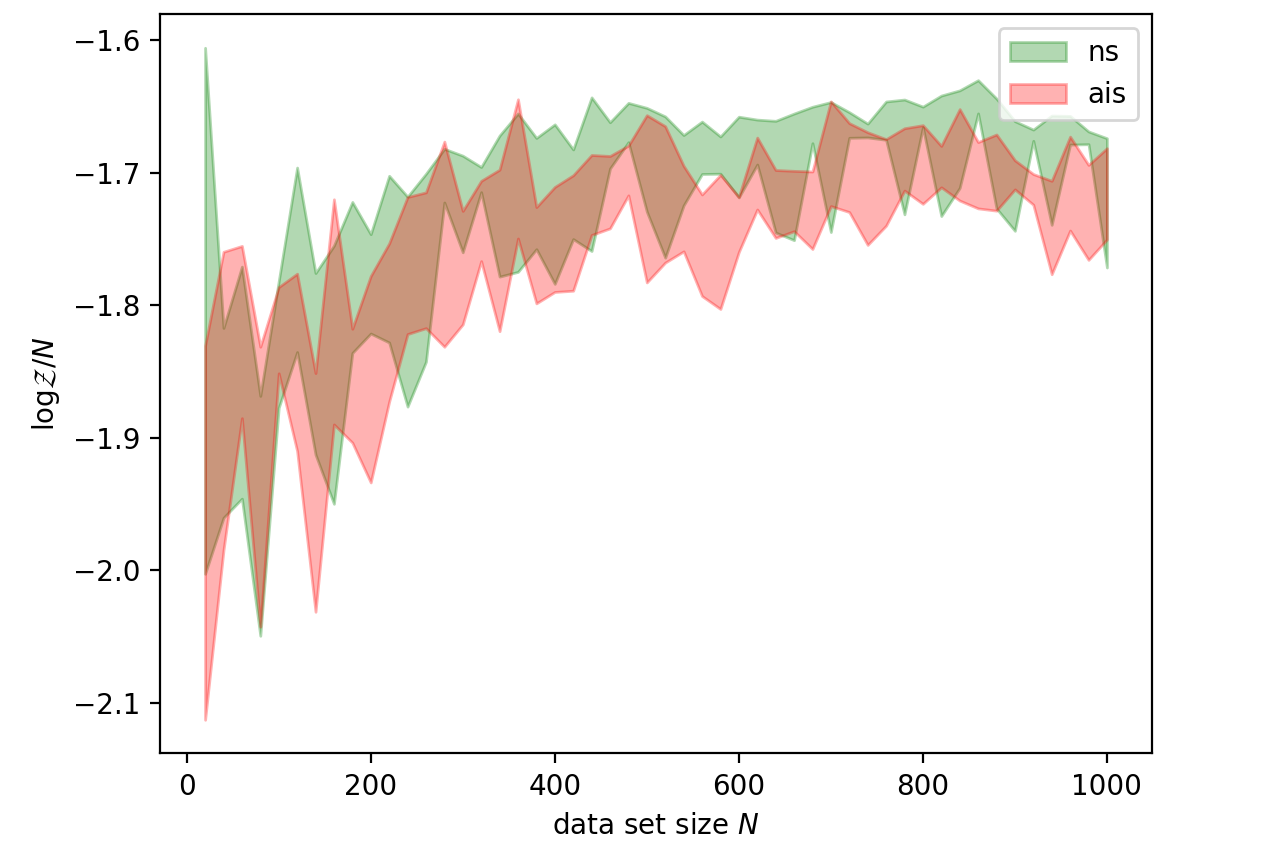}
  \caption{%
    Comparison of NS and AIS with SGHMC kernels for a
    1-dimensional Gaussian mixture model with 9 parameters (8 degrees of
    freedom). The shaded regions are the minimum and maximum of 5 independent
    runs of each algorithm.
  }
  \label{fig:ns-ais}
\end{figure}
}

\section{Model Details\label{ap:models}}
\unskip
\ifdraft{}{
\subsection{Linear Regression}

The data set consists of pairs $(\ubx, y)$ related by
\begin{equation*}
  y = \ubw^T\ubx + b + \epsilon,
\end{equation*}
where $\epsilon$ c distributed with known variance
$\sigma^2$. We do not assume any distribution over $\ubx$ as it always appears
on the right-hand side of the conditional. The single-observation likelihood~is
\begin{equation*}
  p(y|\ubx,\ubw,b) = \frac{1}{\sqrt{2\pi\sigma^2}}\exp\left(-\frac{(y -
      \ubw^T\ubx -
  b)^2}{2\sigma^2}\right).
\end{equation*}
Parameters are $\ubw$ and $b$, with standard Gaussian priors. ML can be
calculated analytically for this model. We used 5 dimensional vectors $\ubx$;
this model therefore has 6 parameters.

\subsection{Logistic Regression}

The data set consists of pairs $(\ubx, y)$, where $\ubx$ is an observation
vector that is assigned a class label $y \in \{1,\ldots,K\}$.  The labels have a
discrete distribution with probabilities given by the softmax function of
an affine transform of the observations
\begin{equation*}
  p(y|\ubx, \bmtheta) = \frac{\exp(\ubw_y^T\ubx + b_y)}{\sum_k \exp(\ubw_k^T\ubx +
  b_k)}.
\end{equation*}
Parameters are $\bmtheta = (\ubw_{1:K}, b_{1:K})$, with standard Gaussian priors.  Again, we do not
assume any distribution over $\ubx$ as it always appears on the right-hand side
of the conditional.
We used 10 dimensional vectors $\ubx$ with 4 classes; this model has 44
parameters.

\subsection{Gaussian Mixture Model}

The data are modeled by a mixture of $d$-dimensional multivariate Gaussian
distributions with diagonal covariance matrices. 
Mixture weights, means, and variances are treated
as parameters. This~type of model is often treated as a latent variable model,
where the mixture component assignments of each data point are the latent
variables. Here, we marginalize out the latent variables to obtain the following
conditional distribution:
\begin{equation*}
  p(\uby|\bmtheta) = \sum_{k=1}^K \beta_k\prod_{j=1}^d
  \frac{1}{\sqrt{2\pi\sigma_{k,j}^2}}
  \exp\left(-\frac{(y_j - \mu_{k,j})^2}{2\sigma_{k,j}^2}\right),
\end{equation*}
\begin{equation*}
  \bmtheta = (\beta_{1:K},\mu_{1:K,1:d},\sigma_{1:K,1:d}^2).
\end{equation*}
Mixture weights $\beta_{1:K}$ are modeled by a Dirichlet prior with $\bm{\alpha}
= 1$; means $\mu_{k,j}$ are modeled conditionally given the variances
by Gaussian priors, centered around zero
with variance $4\sigma_{k,j}^2$; variances $\sigma_{k,j}^2$ are modeled by
inverse gamma priors with shape and scale parameters equal to 1.
We used 5 Gaussian components, and observations were 2-dimensional; this model
has 25 parameters with 24 degrees of freedom.
We use this model for our sensitivity tests, since it has the most complex
structure.
}

\section{Supplementary Figures For Parameter Sensitivity Tests}
\label{ap:figs}
{
\begin{figure}[H]
  \begin{subfigure}{0.5\textwidth}
    \includegraphics[width=\textwidth]{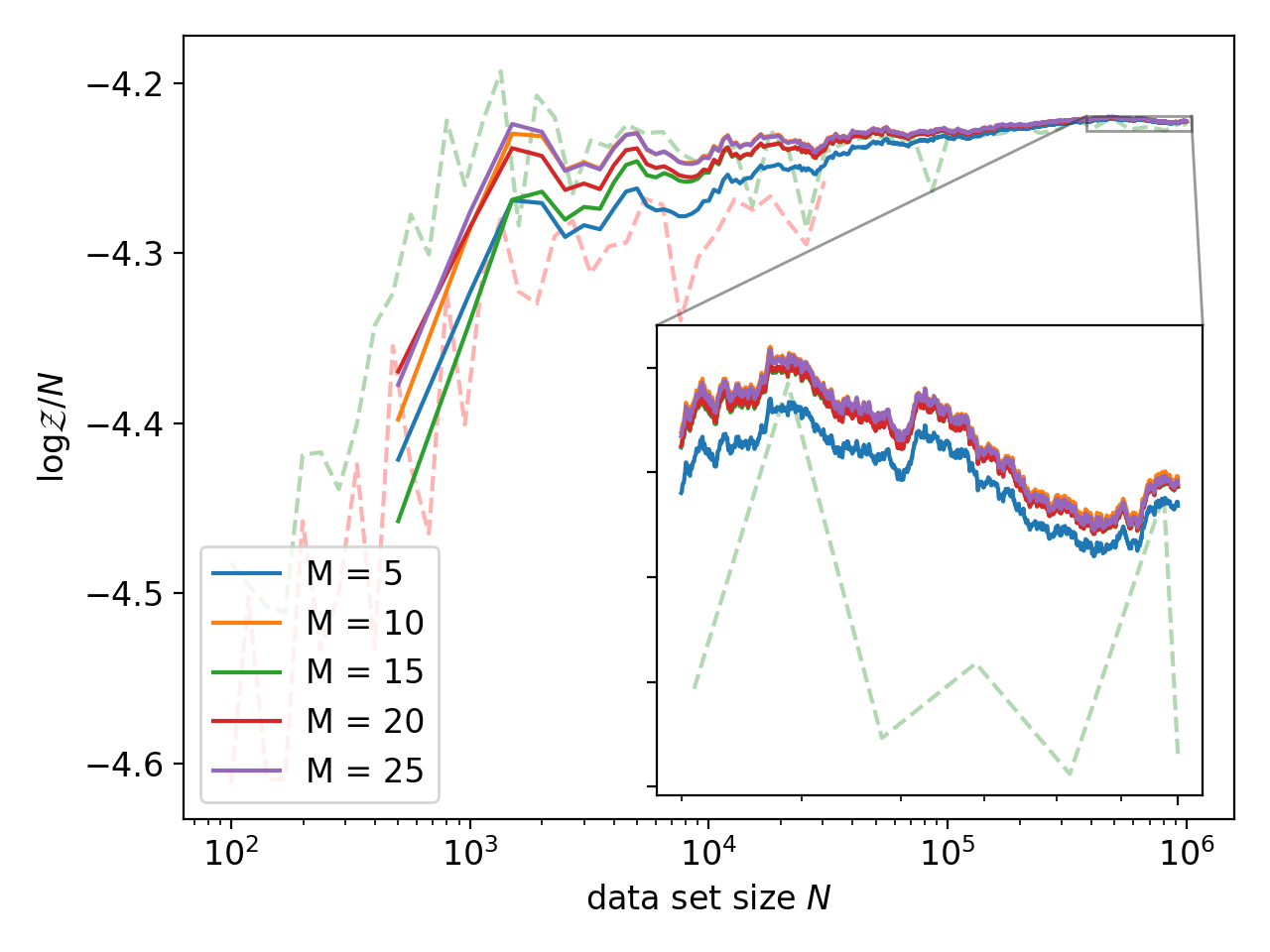}
    \caption{Dependence on $M$}
  \end{subfigure}
  \begin{subfigure}{0.5\textwidth}
    \includegraphics[width=\textwidth]{experiments/figs/gmm-sensitivity-ESS-logz.png}
    \caption{Dependence on $\ESS$}
  \end{subfigure}
  \par\medskip
  \begin{subfigure}{0.5\textwidth}
    \includegraphics[width=\textwidth]{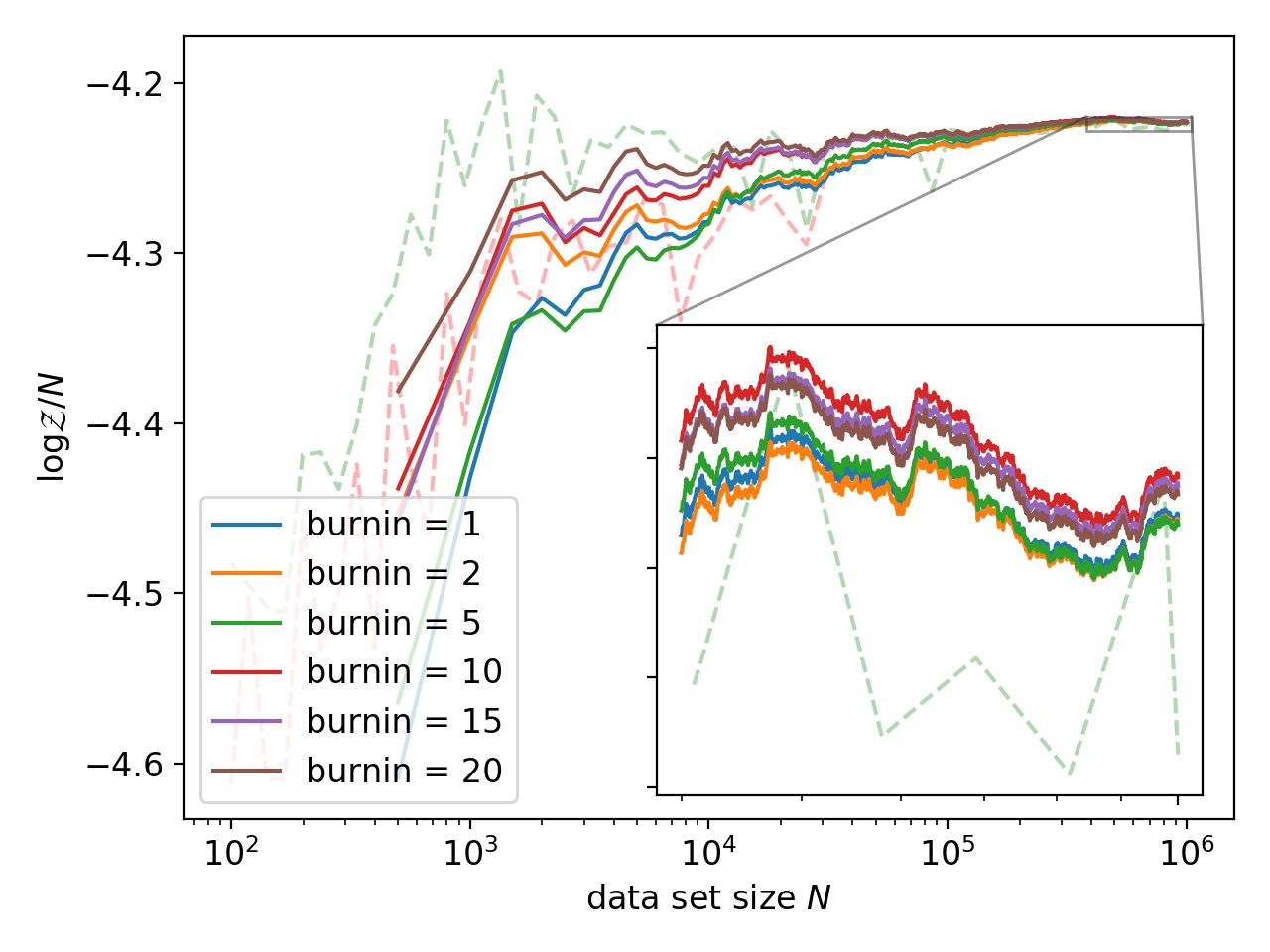}
    \caption{Dependence on number of SGHMC steps}
  \end{subfigure}
  \begin{subfigure}{0.5\textwidth}
    \includegraphics[width=\textwidth]{experiments/figs/gmm-sensitivity-lr-logz.png}
    \caption{Dependence on learning rate}
  \end{subfigure}
  \par\medskip
  \begin{subfigure}{0.5\textwidth}
    \includegraphics[width=\textwidth]{experiments/figs/gmm-sensitivity-lr-burnin-logz.png}
    \caption{Dependence on learning rate with the product of learning rate and
    SGHMC steps constant}
  \end{subfigure}
  \begin{subfigure}{0.5\textwidth}
    \includegraphics[width=\textwidth]{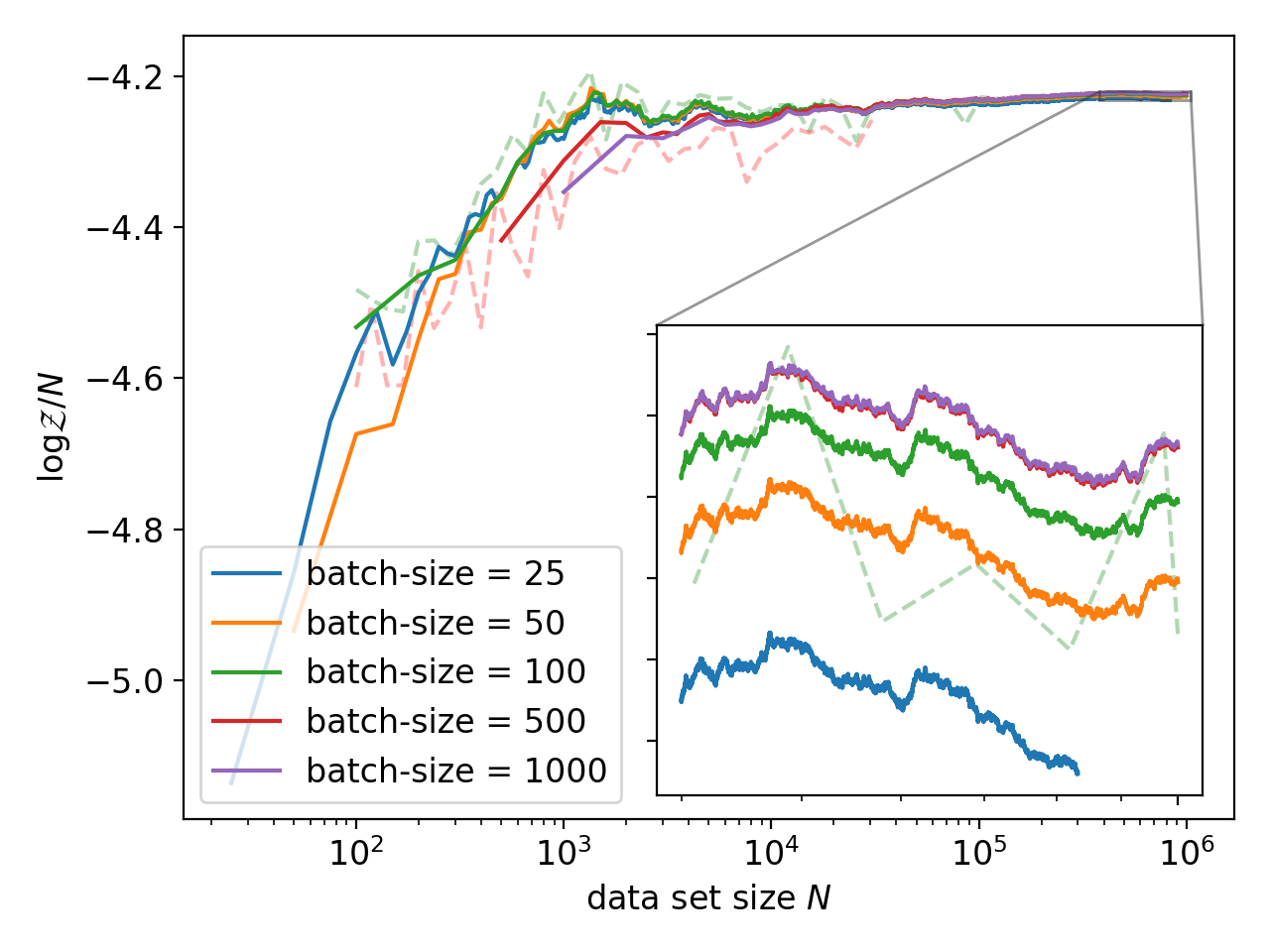}
    \caption{Dependence on mini-batch size.}
  \end{subfigure}
  \caption{Log-ML sensitivity.}
\end{figure}
\unskip
\begin{figure}[H]
  \begin{subfigure}{0.5\textwidth}
    \includegraphics[width=\textwidth]{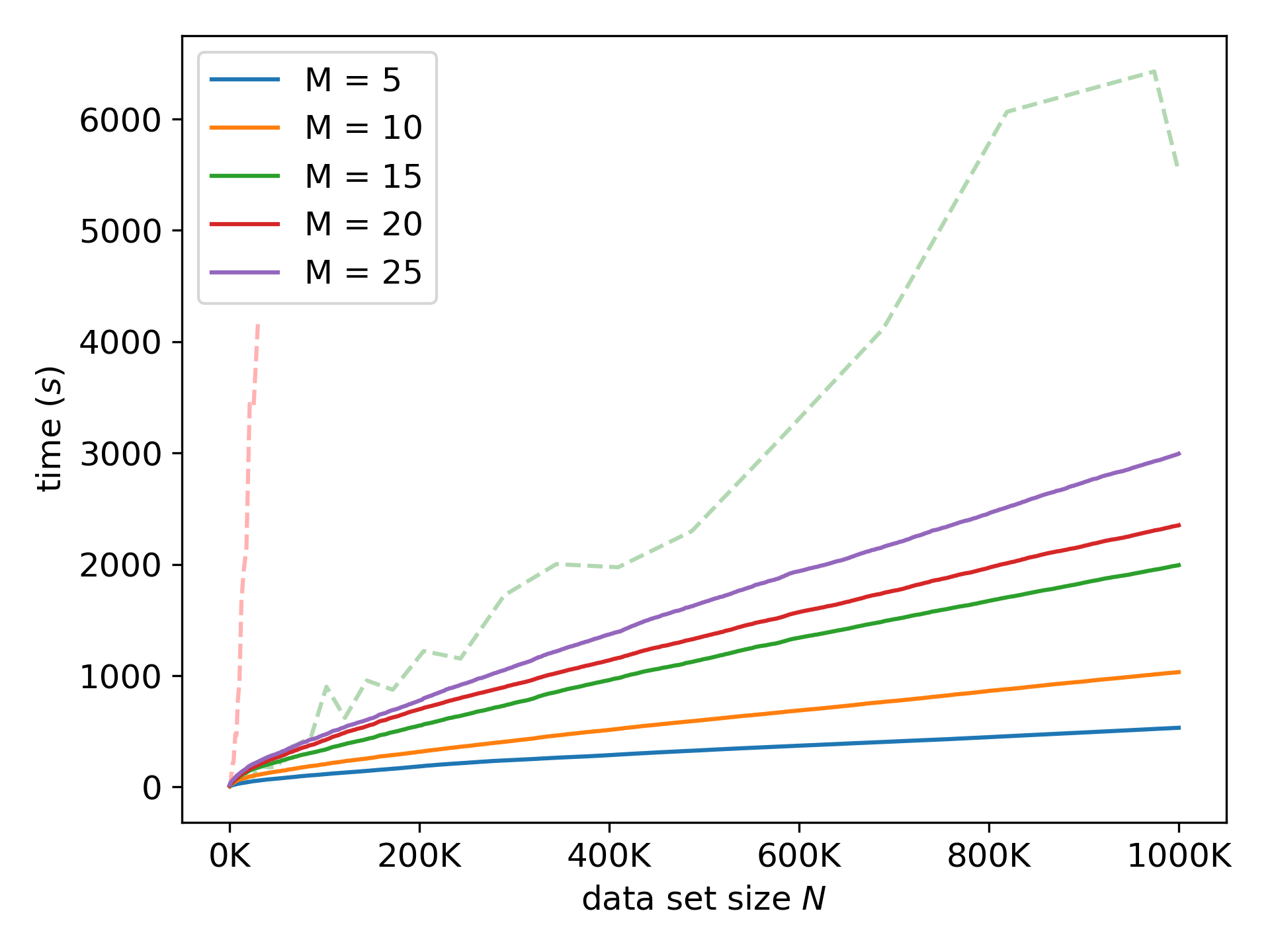}
    \caption{Dependence on $M$}
  \end{subfigure}
  \begin{subfigure}{0.5\textwidth}
    \includegraphics[width=\textwidth]{experiments/figs/gmm-sensitivity-ESS-time.png}
    \caption{Dependence on $\ESS$}
  \end{subfigure}
  \par\medskip
  \begin{subfigure}{0.5\textwidth}
    \includegraphics[width=\textwidth]{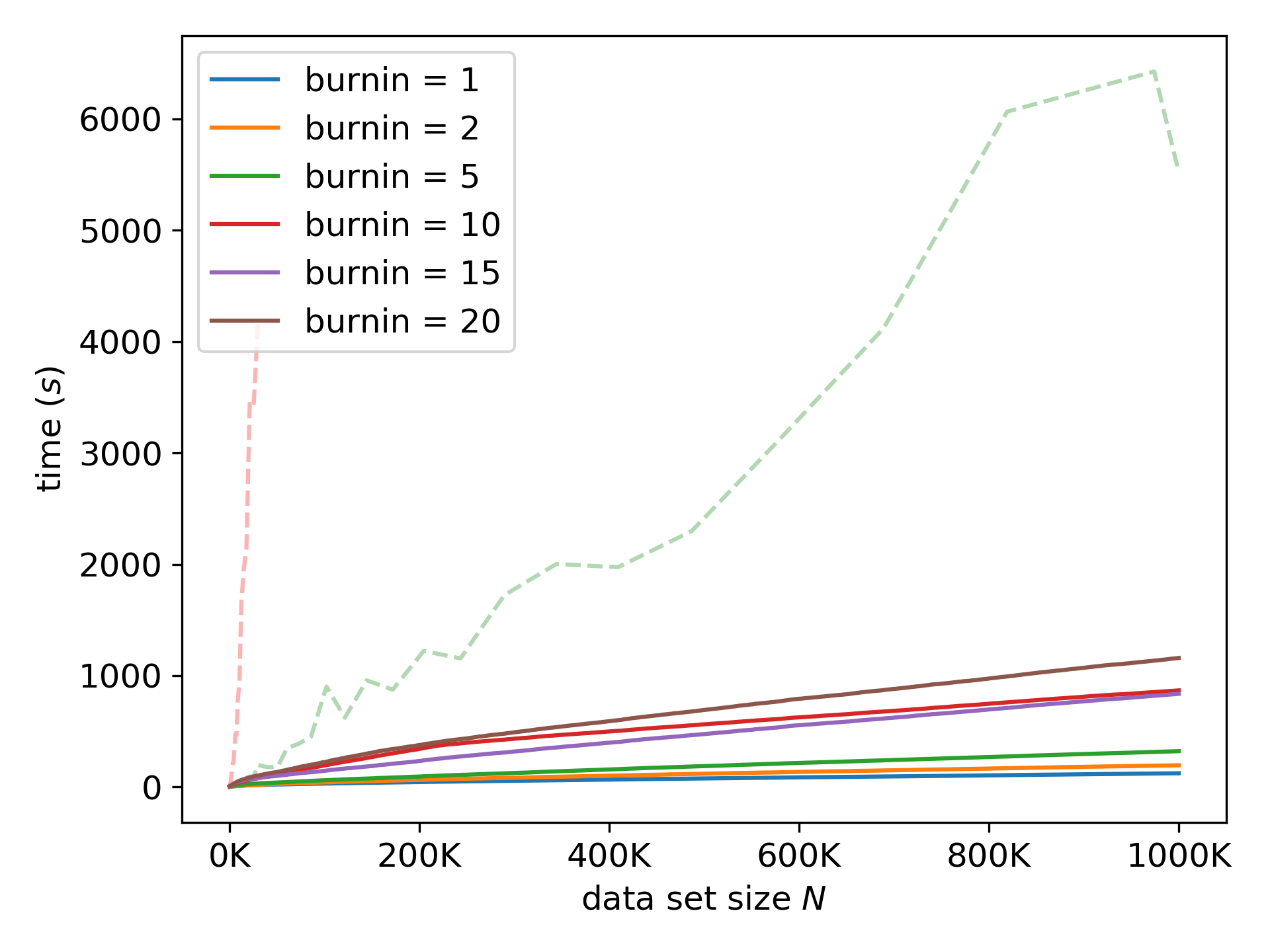}
    \caption{Dependence on number of SGHMC steps}
  \end{subfigure}
  \begin{subfigure}{0.5\textwidth}
    \includegraphics[width=\textwidth]{experiments/figs/gmm-sensitivity-lr-time.png}
    \caption{Dependence on learning rate}
  \end{subfigure}
  \par\medskip
  \begin{subfigure}{0.5\textwidth}
  \centering
    \includegraphics[width=\textwidth]{experiments/figs/gmm-sensitivity-lr-burnin-time.png}
    \caption{Dependence on learning rate with the product of learning rate and
    SGHMC steps constant}
  \end{subfigure}
  \begin{subfigure}{0.5\textwidth}
    \includegraphics[width=\textwidth]{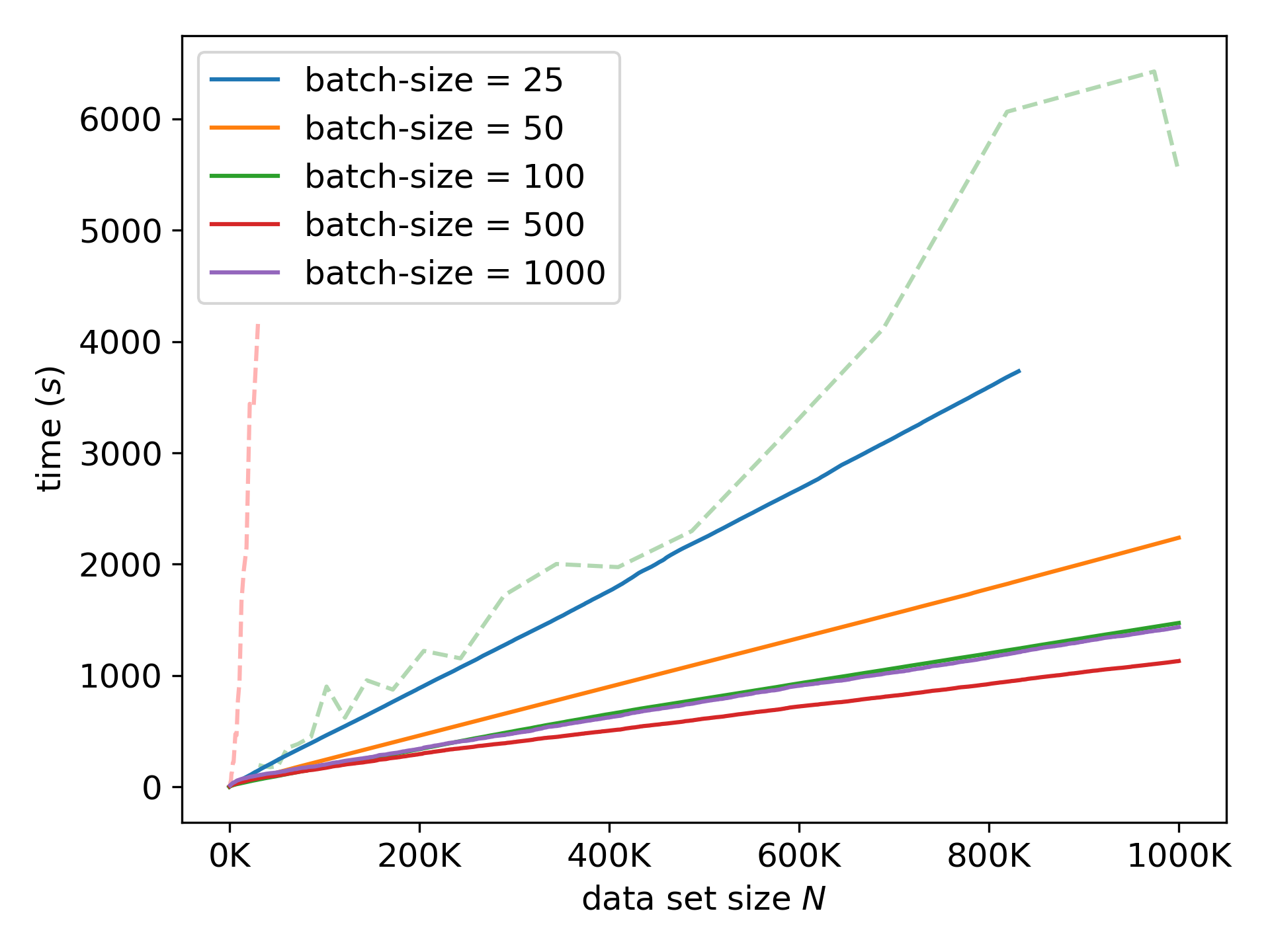}
    \caption{Dependence on mini-batch size}
  \end{subfigure}
  \caption{Running time sensitivity.}
\end{figure}
\unskip
\begin{figure}[H]
  \begin{subfigure}{0.5\textwidth}
    \includegraphics[width=\textwidth]{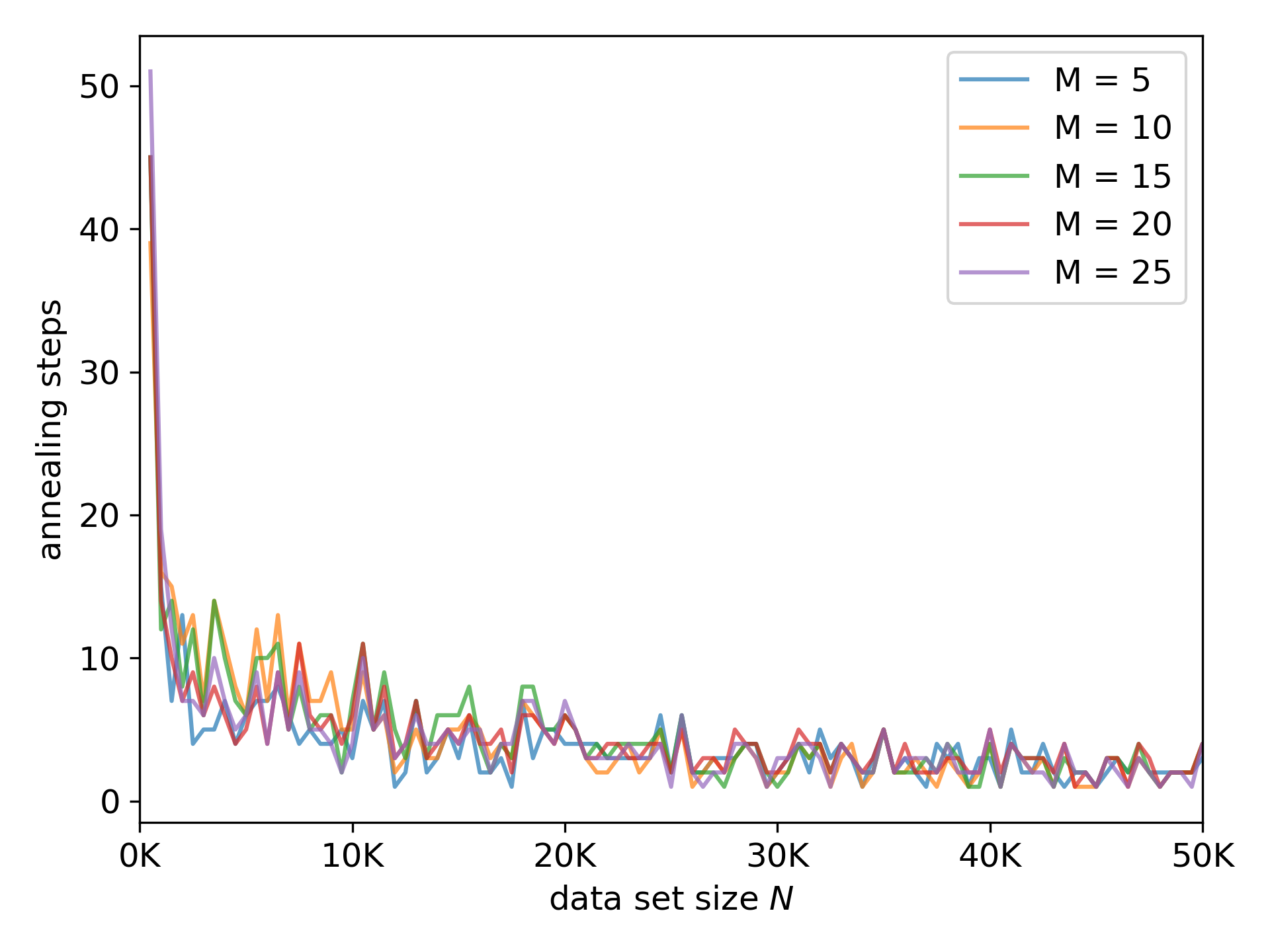}
    \caption{Dependence on $M$}
  \end{subfigure}
  \begin{subfigure}{0.5\textwidth}
    \includegraphics[width=\textwidth]{experiments/figs/gmm-sensitivity-ESS-steps.png}
    \caption{Dependence on $\ESS$}
  \end{subfigure}
  \par\medskip
  \begin{subfigure}{0.5\textwidth}
    \includegraphics[width=\textwidth]{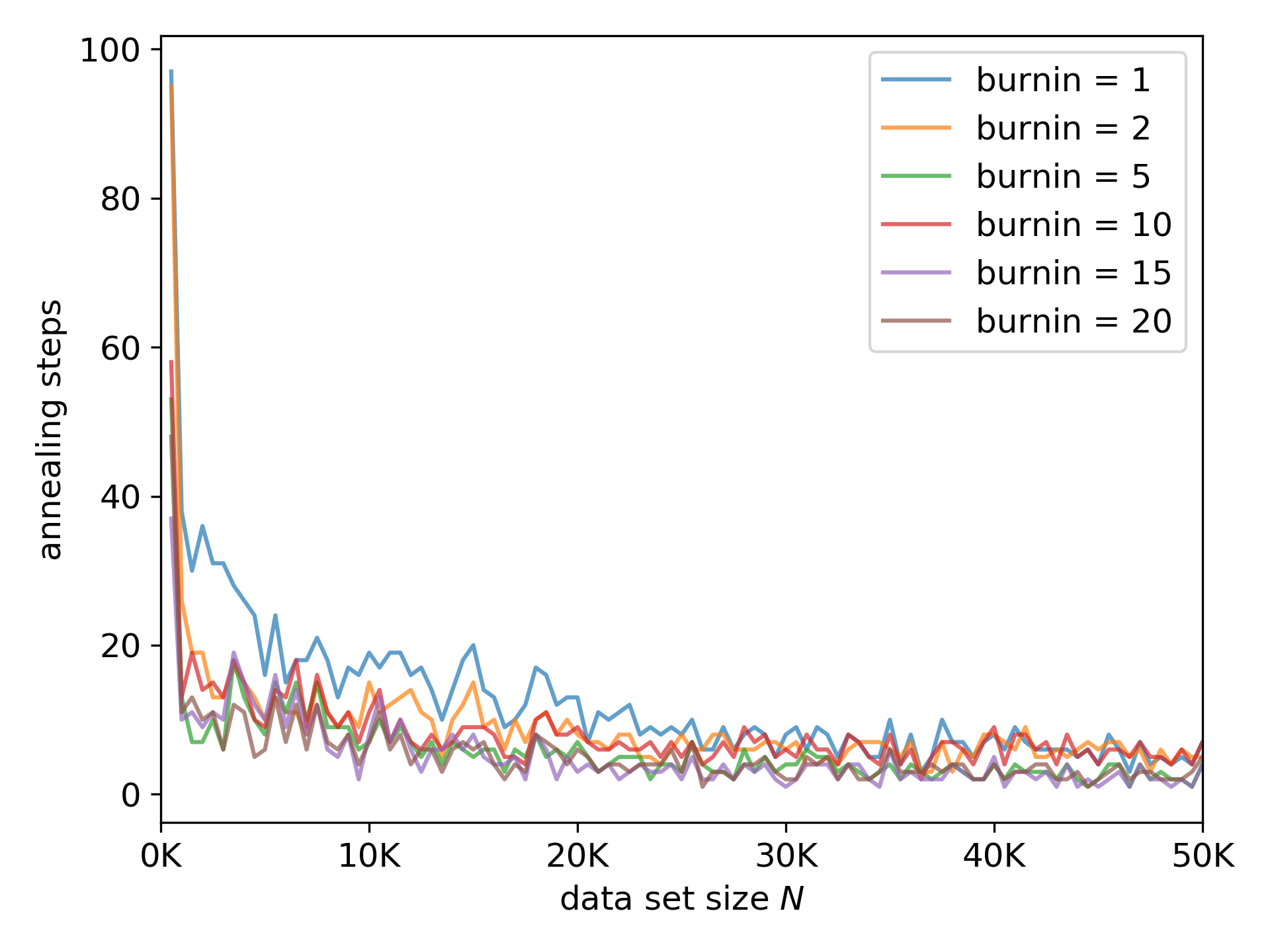}
    \caption{Dependence on number of SGHMC steps}
  \end{subfigure}
  \begin{subfigure}{0.5\textwidth}
    \includegraphics[width=\textwidth]{experiments/figs/gmm-sensitivity-lr-steps.png}
    \caption{Dependence on learning rate}
  \end{subfigure}
  \par\medskip
  \begin{subfigure}{0.5\textwidth}
    \includegraphics[width=\textwidth]{experiments/figs/gmm-sensitivity-lr-burnin-steps.png}
    \caption{Dependence on learning rate with the product of learning rate and
    SGHMC steps constant}
  \end{subfigure}
  \begin{subfigure}{0.5\textwidth}
    \includegraphics[width=\textwidth]{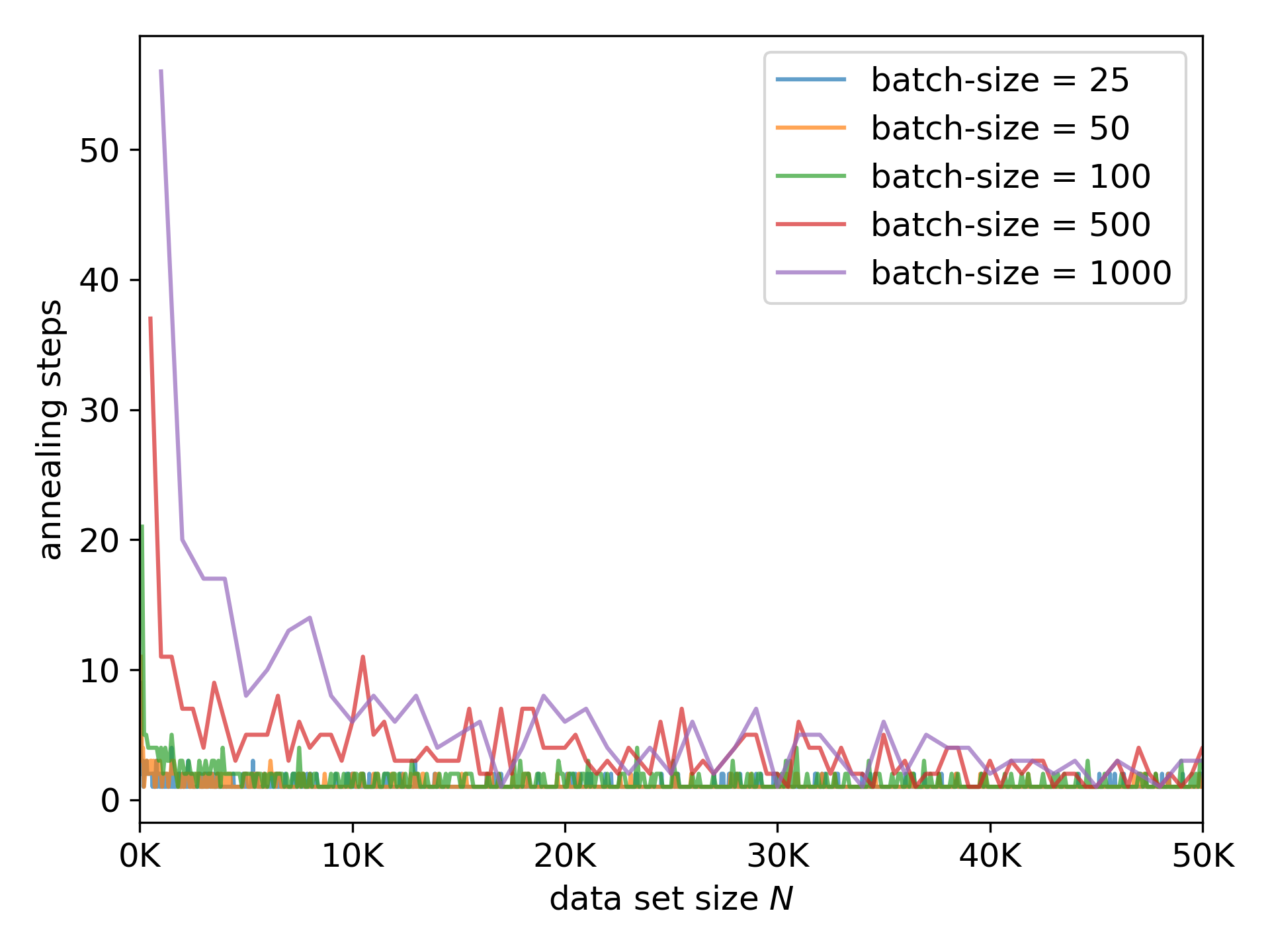}
    \caption{Dependence on mini-batch size}
  \end{subfigure}
  \caption{Sensitivity of number of annealing steps.}
\end{figure}
}

\reftitle{References}


\begin{thebibliography}{999}

\bibitem[Barber(2012)]{bayesian-reasoning}
Barber, D.
\newblock {\em Bayesian Reasoning and Machine Learning}; Cambridge University
  Press: New York, NY, USA,  2012.

\bibitem[Skilling(2006)]{nested-sampling}
Skilling, J.
\newblock Nested Sampling for General {Bayesian} Computation.
\newblock {\em Bayesian Anal.} {\bf 2006}, {\em 1},~833--859,
\newblock
  doi:{\changeurlcolor{black}\href{https://doi.org/10.1214/06-BA127}{\detokenize{10.1214/06-BA127}}}.

\bibitem[{Neal}(1998)]{ais}
{Neal}, R.M.
\newblock {Annealed Importance Sampling}.
\newblock {\em arXiv} {\bf 1998}, arXiv:physics/9803008.

\bibitem[Welling and Teh(2011)]{sgld}
Welling, M.; Teh, Y.W.
\newblock Bayesian Learning via Stochastic Gradient Langevin Dynamics.
\newblock  In Proceedings of the 28th International Conference on Machine
  Learning, {ICML} 2011, Bellevue, WA, USA, 28 June--2 July 2011;
  Getoor, L., Scheffer, T., Eds.; Omnipress: Madison, WI, USA,  2011; pp. 681--688.

\bibitem[{Naesseth} \em{et~al.}(2019){Naesseth}, {Lindsten}, and
  {Sch{\"o}n}]{smc}
{Naesseth}, C.A.; {Lindsten}, F.; {Sch{\"o}n}, T.B.
\newblock {Elements of Sequential Monte Carlo}.
\newblock {\em arXiv} {\bf 2019}, arXiv:1903.04797.

\bibitem[{Gordon} \em{et~al.}(1993){Gordon}, {Salmond}, and
  {Smith}]{bootstrap-filter}
{Gordon}, N.J.; {Salmond}, D.J.; {Smith}, A.F.M.
\newblock Novel Approach to Nonlinear/Non-Gaussian Bayesian State Estimation.
\newblock {\em IEE Proc. F - Radar Signal Process.} {\bf 1993},
  {\em 140},~107--113,
\newblock
  doi:{\changeurlcolor{black}\href{https://doi.org/10.1049/ip-f-2.1993.0015}{\detokenize{10.1049/ip-f-2.1993.0015}}}.

\bibitem[Wallach \em{et~al.}(2009)Wallach, Murray, Salakhutdinov, and
  Mimno]{topic-models}
Wallach, H.M.; Murray, I.; Salakhutdinov, R.; Mimno, D.
\newblock Evaluation Methods for Topic Models.
\newblock  In~Proceedings of the 26th Annual International Conference on Machine
  Learning, Montreal, QC, Canada,  14--18 June 2009; ACM: New York, NY, USA, 2009; pp. 1105--1112,
\newblock
  doi:{\changeurlcolor{black}\href{https://doi.org/10.1145/1553374.1553515}{\detokenize{10.1145/1553374.1553515}}}.

\bibitem[Cameron \em{et~al.}(2019)Cameron, Eggers, and Kroon]{maxent-paper}
Cameron, S.A.; Eggers, H.C.; Kroon, S.
\newblock A Sequential Marginal Likelihood Approximation Using Stochastic
  Graidients.
\newblock  In Proceedings of the 39th International Workshop on Bayesian Inference
  and Maximum Entropy Methods in Science and Engineering, Garching/Munich, Germany, 30 June--5 July 2019.

\bibitem[Chen \em{et~al.}(2014)Chen, B.~Fox, and Guestrin]{sghmc}
Chen, T.; Fox, E.; Guestrin, C.
\newblock Stochastic Gradient {Hamiltonian} {Monte} {Carlo}.
\newblock In Proceedings of the {31st International Conference on Machine Learning,} Beijing, China,  21--26 June
  {2014}; Volume {5}.

\bibitem[{Grosse} \em{et~al.}(2015){Grosse}, {Ghahramani}, and
  {Adams}]{sandwich}
{Grosse}, R.B.; {Ghahramani}, Z.; {Adams}, R.P.
\newblock {Sandwiching the marginal likelihood using bidirectional Monte
  Carlo}.
\newblock {\em arXiv} {\bf 2015}, arXiv:1511.02543.

\bibitem[Kong \em{et~al.}(1994)Kong, Liu, and Wong]{ess}
Kong, A.; Liu, J.S.; Wong, W.H.
\newblock Sequential Imputations and Bayesian Missing Data Problems.
\newblock {\em J. Am. Stat. Assoc.} {\bf 1994},
  {\em 89},~278--288.

\bibitem[{Buchholz} \em{et~al.}(2018){Buchholz}, {Chopin}, and
  {Jacob}]{adaptive-smc}
{Buchholz}, A.; {Chopin}, N.; {Jacob}, P.E.
\newblock {Adaptive Tuning Of Hamiltonian Monte Carlo Within Sequential Monte
  Carlo}.
\newblock {\em arXiv} {\bf 2018}, arXiv:1808.07730.

\bibitem[Beskos \em{et~al.}(2016)Beskos, Jasra, Kantas, and
  Thiery]{adaptive-smc-convergence}
Beskos, A.; Jasra, A.; Kantas, N.; Thiery, A.
\newblock On the Convergence of Adaptive Sequential Monte Carlo Methods.
\newblock {\em Ann. Appl. Probab.} {\bf 2016}, {\em 26},~1111--1146,
\newblock
  doi:{\changeurlcolor{black}\href{https://doi.org/10.1214/15-AAP1113}{\detokenize{10.1214/15-AAP1113}}}.

\bibitem[Durrett(1996)]{stochastic-calculus}
Durrett, R.
\newblock {\em Stochastic Calculus: A Practical Introduction}; Probability and
  Stochastics Series; CRC Press: Boca Raton, FL, USA, 1996;
  pp. 177--207

\bibitem[Gardiner(2004)]{gardiner}
Gardiner, C.W.
\newblock {\em Handbook of Stochastic Methods for Physics, Chemistry and the
  Natural Sciences}, 3rd ed.; {Springer Series in Synergetics};
  Springer: Berlin, Germany, 2004; Volume~13, pp. xviii+415.

\bibitem[{Zhang} \em{et~al.}(2019){Zhang}, {Li}, {Zhang}, {Chen}, and
  {Wilson}]{cyclic-sghmc}
{Zhang}, R.; {Li}, C.; {Zhang}, J.; {Chen}, C.; {Wilson}, A.G.
\newblock {Cyclical Stochastic Gradient MCMC for Bayesian Deep Learning}.
\newblock {\em arXiv} {\bf 2019}, arXiv:1902.03932.

\bibitem[{Ma} \em{et~al.}(2015){Ma}, {Chen}, and {Fox}]{sgrhmc}
{Ma}, Y.A.; {Chen}, T.; {Fox}, E.B.
\newblock {A Complete Recipe for Stochastic Gradient MCMC}.
\newblock {\em arXiv} {\bf 2015},  arXiv:1506.04696.

\bibitem[Springenberg \em{et~al.}(2016)Springenberg, Klein, Falkner, and
  Hutter]{bohamiann}
Springenberg, J.T.; Klein, A.; Falkner, S.; Hutter, F.
\newblock Bayesian Optimization with Robust Bayesian Neural Networks. In {\em
  Advances in Neural Information Processing Systems 29}; Lee, D.D., Sugiyama,
  M., Luxburg, U.V., Guyon, I., Garnett, R., Eds.; Curran Associates, Inc.: Barcelona, Spain
  2016; pp. 4134--4142.

\bibitem[Vitter(1985)]{reservoir-sampling}
Vitter, J.S.
\newblock Random Sampling with a Reservoir.
\newblock {\em ACM Trans. Math. Softw.} {\bf 1985}, {\em 11},~37--57,
\newblock
  doi:{\changeurlcolor{black}\href{https://doi.org/10.1145/3147.3165}{\detokenize{10.1145/3147.3165}}}.

\bibitem[Paszke \em{et~al.}(2017)Paszke, Gross, Chintala, Chanan, Yang, DeVito,
  Lin, Desmaison, Antiga, and Lerer]{pytorch}
Paszke, A.; Gross, S.; Chintala, S.; Chanan, G.; Yang, E.; DeVito, Z.; Lin, Z.;
  Desmaison, A.; Antiga, L.; Lerer,~A.
\newblock Automatic Differentiation in PyTorch. {Available online:
\url{https://openreview.net/pdJsrmfCZ} (accessed on 24 June 2019).}

\bibitem[Skilling(2012)]{gmc-ns}
Skilling, J.
\newblock Bayesian Computation in Big Spaces---Nested Sampling and Galilean
  Monte Carlo.
\newblock {\em AIP Conf. Proc.} {\bf 2012}, {\em 1443},~145--156,
\newblock
  doi:{\changeurlcolor{black}\href{https://doi.org/10.1063/1.3703630}{\detokenize{10.1063/1.3703630}}}.

\end{thebibliography}
\end{document}